\documentclass[12pt]{article}
\usepackage[left=1in, right=1in]{geometry}
\usepackage{amsmath,amssymb,amsthm}
\usepackage{times}
\usepackage{epsfig}
\usepackage{graphicx}
\usepackage{bm}
\usepackage{multirow}
\usepackage{caption}
\usepackage{hhline}
\usepackage{cite}
\usepackage{diagbox}
\usepackage{slashbox,pict2e}
\usepackage{tabu}
\usepackage{mwe}
\usepackage{graphbox}
\usepackage[mathlines]{lineno}
\usepackage{mathtools}
\usepackage{verbatim}
\usepackage{float}
\graphicspath{ {./figure/} }

\usepackage[pagebackref=true,breaklinks=true,letterpaper=true,colorlinks,bookmarks=false]{hyperref}
\usepackage{algorithm,algcompatible,lipsum}
\usepackage{algpseudocode}

\newcommand{\dimin}{N}
\newcommand{\dimout}{M}

\DeclareMathOperator*{\sech}{sech}

\def\note#1{{\color{red}#1}}

\def\l{{\ell}}          
\def\L{{\mathcal{L}}}   
\def\X{{X}}   
\def\Y{{Y}}   
\def\R{\mathbb{R}}

\begin{document}

\title{Adaptive Regularization via Residual Smoothing in Deep Learning Optimization}

\author
{Junghee Cho\thanks{Department of Mathematical Sciences, Seoul National University, Korea} \\  chojunghee@snu.ac.kr \and Junseok Kwon\thanks{Computer Science Department, Chung-Ang University, Korea} \\ jskwon@cau.ac.kr \and Byung-Woo Hong\footnotemark[2] \\ hong@cau.ac.kr
}
\maketitle

\begin{abstract}
   We present an adaptive regularization algorithm that can be effectively applied to the optimization problem in deep learning framework.
   Our regularization algorithm aims to take into account the fitness of data to the current state of model in the determination of regularity to achieve better generalization.
   The degree of regularization at each element in the target space of the neural network architecture is determined based on the residual at each optimization iteration in an adaptive way.
   Our adaptive regularization algorithm is designed to apply a diffusion process driven by the heat equation with spatially varying diffusivity depending on the probability density function following a certain distribution of residual. 
   Our data-driven regularity is imposed by adaptively smoothing a simplified objective function in which the explicit regularization term is omitted in an alternating manner between the evaluation of residual and the determination of the degree of its regularity.
   The effectiveness of our algorithm is empirically demonstrated by the numerical experiments in the application of image classification problems, indicating that our algorithm outperforms other commonly used optimization algorithms in terms of generalization using popular deep learning models and benchmark datasets.
\end{abstract}


%
%
\section{Introduction}
%
%
Deep neural networks have made a significant progress in a variety of applications at a number of domains such as image understanding~\cite{cai2018learning}, sound recognition~\cite{mesaros2018detection}, motion planning~\cite{shahroudy2018deep}, and other decision support~\cite{Ahmad2019EventRecog}.
In particular, the successful application of convolutional neural networks (CNNs)~\cite{lenet} to the computer vision problems has driven advanced performance in a variety of applications such as recognition~\cite{szegedy2017inception}, segmentation~\cite{chen2018deeplab}, motion estimation~\cite{ren2017unsupervised} or reconstruction~\cite{zhang2017beyond} due to their effective characteristic power and generalization capabilities, leading to large scale optimization problems where the numbers of both model parameters and training data are often huge.
The optimization in the deep learning applications often involves the stochastic estimation of gradients using the stochastic gradient descent in order to improve the computational efficiency with a large number of training data.
Albeit the choices of mini-batch size and learning rate are implicitly related to the generalization of the model~\cite{levy2017online}, it is generally required to introduce an explicit regularization term in the objective function to avoid over-parameterization or over-fitting.  
The objective function mainly consists of a data fidelity term that measures a discrepancy between estimation and observation and a regularization term that imposes smoothness constraint in the solution space, and their relative significance is usually determined by a constant based on the ratio of variances between likelihood and prior distributions.
However, the computation of those distributions is computationally intractable leading to the grid search approach in determining the control parameter between the data fidelity and the regularization. 
In addition, the choice of static control parameter implies that the underlying likelihood and prior probabilities follow single model distributions, which is often undesirable to represent complicated models.
%
%
%
%
%
    
In this work, we propose a simple, yet effective regularization scheme that is designed to impose adaptive regularity depending on both spatial and temporal domain of optimization.
We consider residual that is indicative of fitness between data and the current state of model in the determination of regularization in such a way that the adaptive application of regularization is achieved in both space and time for better generalization..
We develop an implicit regularization scheme based on a simplified objective function where the regularization term is omitted and a diffusion process is applied to the data fidelity term.
The diffusivity of diffusion process driven by heat equation is determined based on a probability density function following a certain distribution of residual at each residual element in the course of optimization.
%
In the application of our approach to the deep learning algorithm, we present a neural network architecture incorporating our adaptive regularization, which is efficiently implemented by an additional smoothing layer with a deterministic smoothing kernels.
%
We present the effectiveness of our proposed algorithm for generalization of model in the application of image classification problems with popular network models and commonly used benchmark datasets while our algorithm can be naturally integrated with other architectures of networks such as autoencoder for image segmentation or motion estimation.  

In the remainder of this paper, we relate our method to the prior works in Sec.~\ref{sec:related} and present the conventional optimization algorithm in Sec.~\ref{sec:prelimminary} followed by our proposed algorithm in Sec.~\ref{sec:residual}. 
The implementation of our adaptive regularization algorithm in the deep neural network framework is provided in Sec.~\ref{sec:netwrok} and the results of numerical experiments are presented in Sec.~\ref{sec:exp} and the conclusion follows in Sec.~\ref{sec:conclusion}.
%
%
\section{Related Work} \label{sec:related}
There have been a variety of regularization techniques in machine learning applications. 
One can categorize the techniques into two classes, namely, explicit regularization and implicit regularization.
We provide a number of algorithms for the explicit regularization and the implicit regularization in Sec.~\ref{sec:explicit} and Sec.~\ref{sec:implicit}, respectively.
Then, we discuss in more detail the closely related works to our algorithm in Sec.~\ref{sec:smoothing} where the smoothing technique is considered to impose regularity on the solution space.

\subsection{Explicit Regularization} \label{sec:explicit}
\noindent{\bf Weight Decay: } 
The objective function is assumed to include a regularization term that penalizes a perturbation of unknown parameters in terms of $L_2^2$ norm.
The gradient descent of the regularization term yields the decay of weights in a recursively manner with a given rate parameter and a learning rate. 
It is considered as one of the most practical regularization algorithms due to its computational convenience, yet often blur the solution.

\noindent{\bf Sparsity Constraint: } 
Sparsity has emerged as a way to impose $L_1$ regularization to objective functions.
The essential motivation of the sparsity assumption on the solution space stems from the modeling of the residual distribution with a sharp peak, which is known to be more realistic in most real-world problems.
Sparsity constraints suppress undesirable perturbations while preserving discontinuities in order to avoid over-fitting.

\noindent{\bf Entropy Minimization: } 
In the application of sparsity constraint to the probability distribution of solution, the entropy term in the objective function has been introduced in~\cite{Grandvalet2004Entropy} where the entropy is to be minimized. Entropy minimization has been shown to improve exploration ability, thus can regularize the objective functions in reinforcement learning tasks~\cite{Jordi2019Entropy}.

In contrast to the above explicit regularization techniques, our algorithm bases on the objective function that omits the regularization term instead applying simple, yet effective diffusion process to the data fidelity term.
\subsection{Implicit Regularization} \label{sec:implicit}
\noindent{\bf Noise Injection: } 
In the estimation of gradients using the stochastic gradient descent, stochastic noise is involved and its variance is related to the size of mini-batch. The injection of stochastic noise to the neural network can be used as a way to impose regularization to arrive at a better local minimum~\cite{kingma2015variational}.
It is also shown that the variance of injected noise is related to the amount of imposed smoothness on the solution~\cite{noh2017regularizing} where a tighter lower bounds of the objective function can be achieved by adding noises in a stochastic gradient descent iteration. 
In addition to the manipulation of noise, smoothing of ground truth label has been proposed to make the model less confident regarding its trained weights, thus improving generality~\cite{szegedy2016rethinking} where the probability of each label is arbitrarily perturbed depending on a random distribution.
Similarly, there has been a regularization algorithm that replaces one of the ground truth labels with an arbitrary label uniformly at random~\cite{xie2016disturblabel}. 

\noindent{\bf Dropout: } 
One of the implicit implementations of sparsity constraints that suppress the value of weights to be zero is Dropout~\cite{srivastava2014dropout} that randomly eliminates units of the neural network with a uniform probability while training, thus prevents units from excessive co-adapting.
There have been a variety of Dropout techniques including maxout network~\cite{goodfellow2013maxout} that proposes a new activation function to leverage the dropout, stochastic pooling~\cite{zeiler2013stochastic} that replaces the deterministic pooling with a stochastic procedure by randomly choosing activation from a multinomial distribution.

\noindent{\bf Learning Rate Decay: } 
The stochastic gradient descent often yields better training results with a learning rate annealing scheme that schedules a temporal series of learning rates in epoch where a decreasing scheduling is generally applied to improve convergence.
Whereas, a decreasing annealing pattern has been repeated as a warm start to overcome undesirable sharp local minima in~\cite{loshchilov2016sgdr}.

\noindent{\bf Model Ensemble: }
There has been a regularization technique developed by combining differently trained neural networks and introducing regularization effects imposed by different network architectures~\cite{singh2016swapout}.
The random ensemble of prediction functions is known to provide better training behavior~\cite{Xuanqing2018} and the structural dropout, called Branchout~\cite{Han2017Branchout}, has been developed by randomly choosing a subset of branches in the convolutional neural networks.

\noindent{\bf Batch Normalization: } 
Batch normalization~\cite{ioffe2015batch} has been proposed for resolving the internal covariant shift by normalizing layer inputs, in which the distribution of inputs of each layer changes during the training process.
However, batch normalization is proven that it can also improve the regularization performance in training neural networks~\cite{Luo2019}.
In addition, batch normalization enabled training with larger learning rates, which induces faster convergence and better generalization~\cite{Bjorck2018}.

While the aforementioned implicit methods mainly impose global regularity on the solution space, our method uses the residual that is variable in energy space and optimization time, thus spatially and temporally varying regularization depending on the residual.
\subsection{Regularization via Energy Smoothing} \label{sec:smoothing}
There is different perspective of imposing regularity that the geometric property of energy landscape is modified in such a way that undesirable insignificant local minima are eliminated by smoothing the energy~\cite{mobahi2015link} where the objective function is convolved with Gaussian kernels.
The approximated solution to the specific evolutionary partial differential equation (PDE) leads to convex envelopes of the objective function, but the approximation is assumed to be a solution of the PDE with small perturbations, which is often not the case. A modified network has been proposed in~\cite{gulcehre2016mollifying} where the loss function is differentiable, smooth and computationally stable.

Unlike the conventional smoothing approaches for regularization in deep learning optimization, our algorithm considers diffusion of residual with spatially and temporally varying diffusivity leading to adaptive regularization that is more suited for complex models in a variety of deep learning applications. 
%

%
%
\section{Preliminary} \label{sec:prelimminary}
We consider a minimization problem in a supervised learning framework. Let $\chi = \{ (x_i, y_i) \}_{i = 1}^n$ be a set of training data where $x_i \in \X \subset \mathbb{R}^\dimin$ is the $i$-th input and $y_i \in \Y \subset \mathbb{R}^\dimout$ is its desired output. 
Let $h_w \colon \X \to \Y$ be a prediction function that is associated with its model parameters $w = ( w_1, w_2, \cdots, w_m ) \in \R^m$ where the dimension of the feature space is $m$.
The objective of the supervised learning problem is to find optimal parameters $w^*$ that are typically obtained by minimizing the empirical loss $\L(w)$ defined on the training data $\chi$:
\begin{linenomath*}
    \begin{align}
    \L(w) &= \frac{1}{n} \sum_{i=1}^n f_i( w ) + \lambda \, \gamma( w ), \label{eq:objective}
    \end{align}
\end{linenomath*}
where we denote by $f_i(w)$ a data fidelity term for a pair of data $(x_i, y_i)$ and by $\gamma(w)$ a regularization term, and $\lambda > 0$ is a control parameter for the balance between the two terms.
%
The data fidelity $f_i(w)$ incurred by a set of parameters $w$ with a sample $(x_i, y_i)$ is designed to measure the discrepancy between the prediction $h_w(x_i)$ with input $x_i$ and its desired output $y_i$. 
The regularization $\gamma(w)$ aims to impose smoothness condition on the prediction function $h_w(x_i)$, thus avoid over-fitting of the model.
The control parameter $\lambda$ is determined based on the relation between the underlying distribution of data and the prior distribution of model.

We consider a first-order optimization algorithm to minimize the objective function that is assumed to be differentiable leading to the following gradient descent step at iteration $t$:
\begin{linenomath*}
    \begin{align}
    w^{t+1} \coloneqq w^{t} - \eta^{t} \left( \frac{1}{n} \sum_{i=1}^n \nabla f_{i}(w^{t}) + \lambda \nabla \gamma(w^t) \right), \label{eq:update:vanilla}
    \end{align}
\end{linenomath*}
where we denote by $\nabla f_{i}(w^{t})$ gradient of $f_i$ with respect to $w$ at iteration $t$, and by $\eta^{t}$ the learning rate.  
The computation of the above full gradient over the entire training data is often intractable due to a large number of data, which leads to the use of stochastic gradient that is computed using a subset uniformly selected at random from the training data.
The iterative step of the stochastic gradient descent algorithm at iteration $t$ reads:
\begin{linenomath*}
    \begin{align}
    w^{t+1} \coloneqq w^{t} - \eta^{t} \left( \frac{1}{|\beta^t|} \sum_{i \in \beta^t} \nabla f_{i}(w^{t}) + \lambda \nabla \gamma(w^t) \right), \label{eq:update:std}
    \end{align}
\end{linenomath*}
where $\beta^t$ denotes a mini-batch that is the index set of a subset uniformly selected at random from the training data. The size of mini-batch $|\beta^t|$ is related to the variance of the gradient norms, and thus to the regularization of the model. The small size of mini-batch yields stochastic gradients with higher variance due to noise involved in the stochastic process leading to large regularization.   
%
%
\section{Regularization via Residual Smoothing} \label{sec:residual}
The optimization of interest aims to minimize the objective function that consists of a data fidelity term, a regularization term, and a control parameter for their relative weight.
The selection of control parameter is often critical to obtain a better solution and is determined by the ratio between the underlying distributions of the residual and the prior smoothness, both of which are mostly assumed to follow unimodal distributions.
Thus, the control parameter is chosen to be constant, and it is generally required to apply a grid search over a range of parameters to choose optimal parameters.
However, it is often ineffective to model the distribution of data fidelity and determine the ratio of its variance to the variance of prior distribution for a smooth solution based on a unimodal probability density function leading to a static control parameter for the trade-off between data fitting and smoothness.
Thus, we propose an adaptive regularization scheme that considers residual in the determination of regularity at each point of the residual domain.  
%
%
\subsection{Adaptive Regularization based on Residual} 
The computation of empirical stochastic gradient involves the noise process following a certain distribution with zero mean, and its variance is inversely proportional to the size of mini-batch. In addition to the stability, the noise process is also related to the regularization, thus the size of mini-batch can be used in determining regularity in an implicit way.
On the other hand, the control parameter $\lambda$ in the objective function in~\eqref{eq:objective} can be variable with a fixed mini-batch size for each sample $(x_i, y_i)$, leading to the following modified objective function:
\begin{linenomath*}
    \begin{align}
    \tilde{\L}(w) &= \frac{1}{n} \sum_{i=1}^n \left( f_i( w ) + \lambda_i \, \gamma( w ) \right), \label{eq:objective:adaptive}
    \end{align}
\end{linenomath*}
where $\lambda_i \in \R$ denotes a weighting parameter for the regularization term and it is designed to be associated with each sample $(x_i, y_i)$.
We assume that the degree of regularity follows a distribution of the residual leading to the following data-driven regularity:
\begin{linenomath*}
    \begin{align}
    \lambda_i \propto 1 - \exp\left( - \frac{\| f_i(w) \|}{\nu} \right), \label{eq:adaptive:distribution}
    \end{align}
\end{linenomath*}
where $\nu$ is a parameter for the variance of the residual.
The degree of regularity is designed to be proportional to the magnitude of residual for each sample.
In addition to the adaptive application of regularity with respect to sample, we consider the temporal state of solution in the course of optimization leading to the update of model parameters based on the stochastic gradients incorporating data-driven regularity as follows:
\begin{linenomath*}
    \begin{align}
    w^{t+1} \coloneqq w^{t} - \eta^{t} \frac{1}{|\beta^t|} \sum_{i \in \beta^t} \left( \nabla f_i(w^t) + \lambda_i^t \nabla \gamma(w^t) \right), \label{eq:update:adaptive}
    \end{align}
\end{linenomath*}
where $\lambda_i^t$ is variable with respect to both optimization iteration $t$ and sample index $i$.
The intrinsic motivation of the temporally adaptive regularization stems from the limitation of the existing static scheme that imposes the same degree of regularity albeit the residual decays in the optimization steps.
However, it is computationally expensive to construct the distribution from which the control parameter for regularization is determined while computing the gradients of both data fidelity and regularization terms.
In addition to the computational efficiency, it is desired to consider the relative magnitude of residual in its spatial domain.
Thus, we propose a simple, yet effective regularization scheme that is designed to impose adaptive regularity depending on both spatial and temporal domain of optimization, which is achieved by smoothing residual with spatially and temporally varying degree without explicit computation of gradient for the regularization term that is omitted in the objective function, as presented in the following section.
%
%
\subsection{Regularization via diffusion process that is adaptive in the both spatial and temporal domains}
We propose a regularization algorithm which is developed based on smoothing the residual that measures the discrepancy between model and sample data without taking into account an explicit regularization term.
We modify the objective function in Eq.~\eqref{eq:objective:adaptive} from which the regularization term is omitted and the original data fidelity term $f_i(w)$ is replaced with $g_i(w)$ as follows:
\begin{linenomath*}
    \begin{align}
    \bar{\L}(w) &= \frac{1}{n} \sum_{i=1}^n g_i(w), \quad g_i(w) = \| u_i(w) \|_2^2, \label{eq:objective:smoothing}
    \end{align}
\end{linenomath*}
where $g_i(w)$ is $L_2^2$ norm of the diffused residual $u_i(w)$ for each sample $(x_i, y_i)$, and $u_i(w)$ is obtained by imitating the diffusion process using the heat equation as follows:
\begin{linenomath*}
    \begin{align}
        \begin{cases}
            \frac{\partial u_i(w; \tau)}{\partial \tau} &= \kappa \, \Delta u_i(w; \tau),\\
            u_i(w; 0) &= d_i(w), \label{eq:heat}
        \end{cases}
    \end{align}
\end{linenomath*}
where $\kappa$ denotes a diffusion coefficient, $\Delta$ the Laplace operator, and $\tau$ an auxiliary variable for the diffusion time. When $\kappa$ is constant, the solution of heat equation is given by the convolution of initial data with the Gaussian kernel and it implies that the value at every point of solution becomes gradually averaged from its surrounding data as the diffusion proceeds. The Neumann boundary condition is imposed and the initial condition $u_i(w; 0)$ is given by the residual defined by the magnitude of the discrepancy between the predication and the desired output as follows:
\begin{linenomath*}
    \begin{align}
    d_i(w) &= | h_w(x_i) - y_i |, \label{eq:residual}
    \end{align}
\end{linenomath*}
where $d_i(w) \in \R^{\dimout}$.
In the diffusion equation, the coefficient $\kappa$ is normally set to be constant, but we consider a diffusivity map $\kappa \colon \R^\dimout \to \R^\dimout$ that is employed to impose spatially varying regularity depending on the residual.
The diffusivity map is designed to apply regularity following a distribution of residual based on the sigmoid function $S(x; s, \alpha)$ defined by:
%
\begin{linenomath*}
    \begin{align}
    S(x; s, \alpha) = \frac{s}{1 + \exp(- \alpha x)}, \label{eq:sigmoid}
    \end{align}
\end{linenomath*}
where $s, \alpha \in \R^+$ are parameters that determine the vertical scale and the steepness of transition in function value, respectively. 

\begin{figure}[htb]
\centering
\begin{tabular}{c@{\hspace{10pt}}c}
\includegraphics[height=80pt]{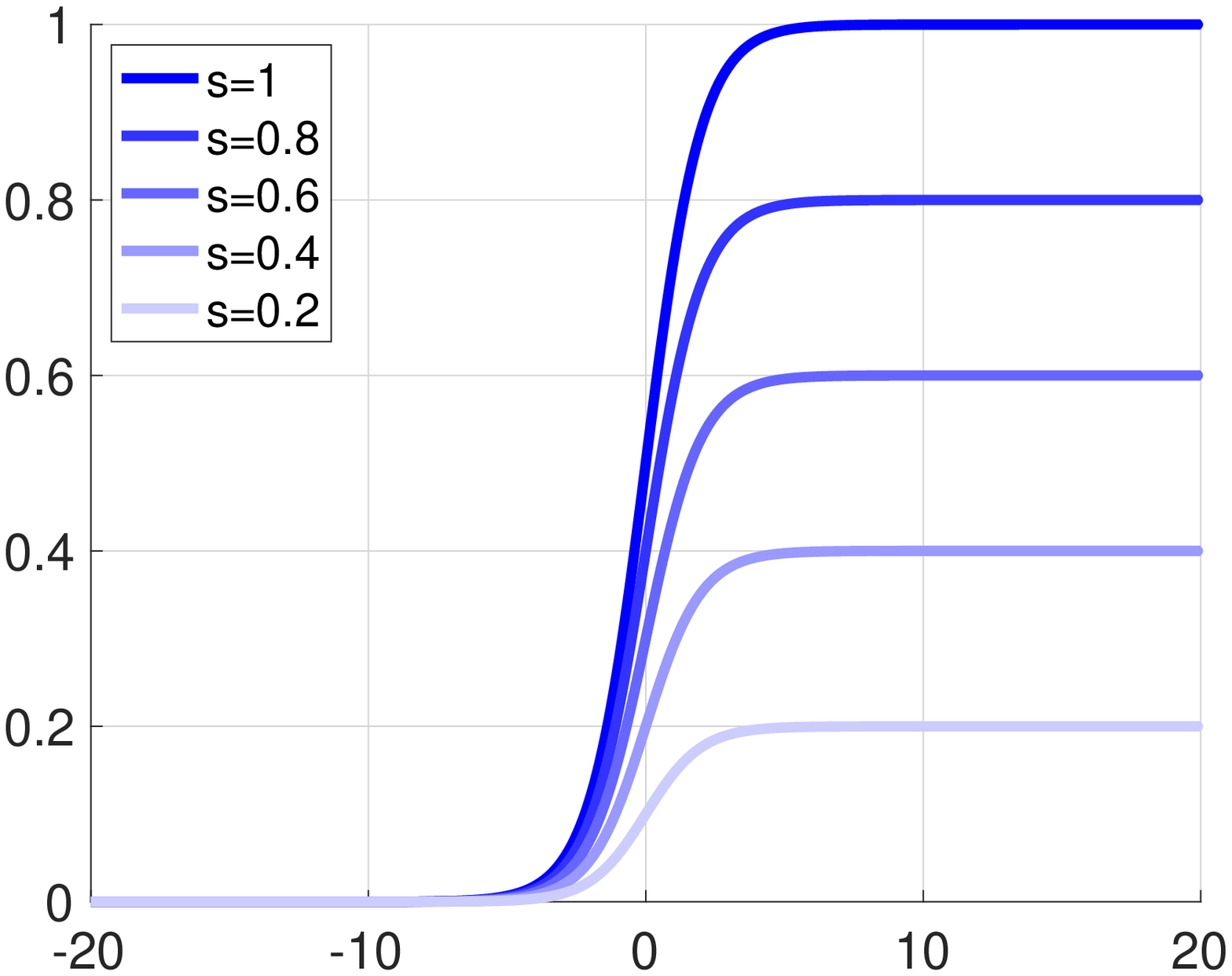}
& 
\includegraphics[height=80pt]{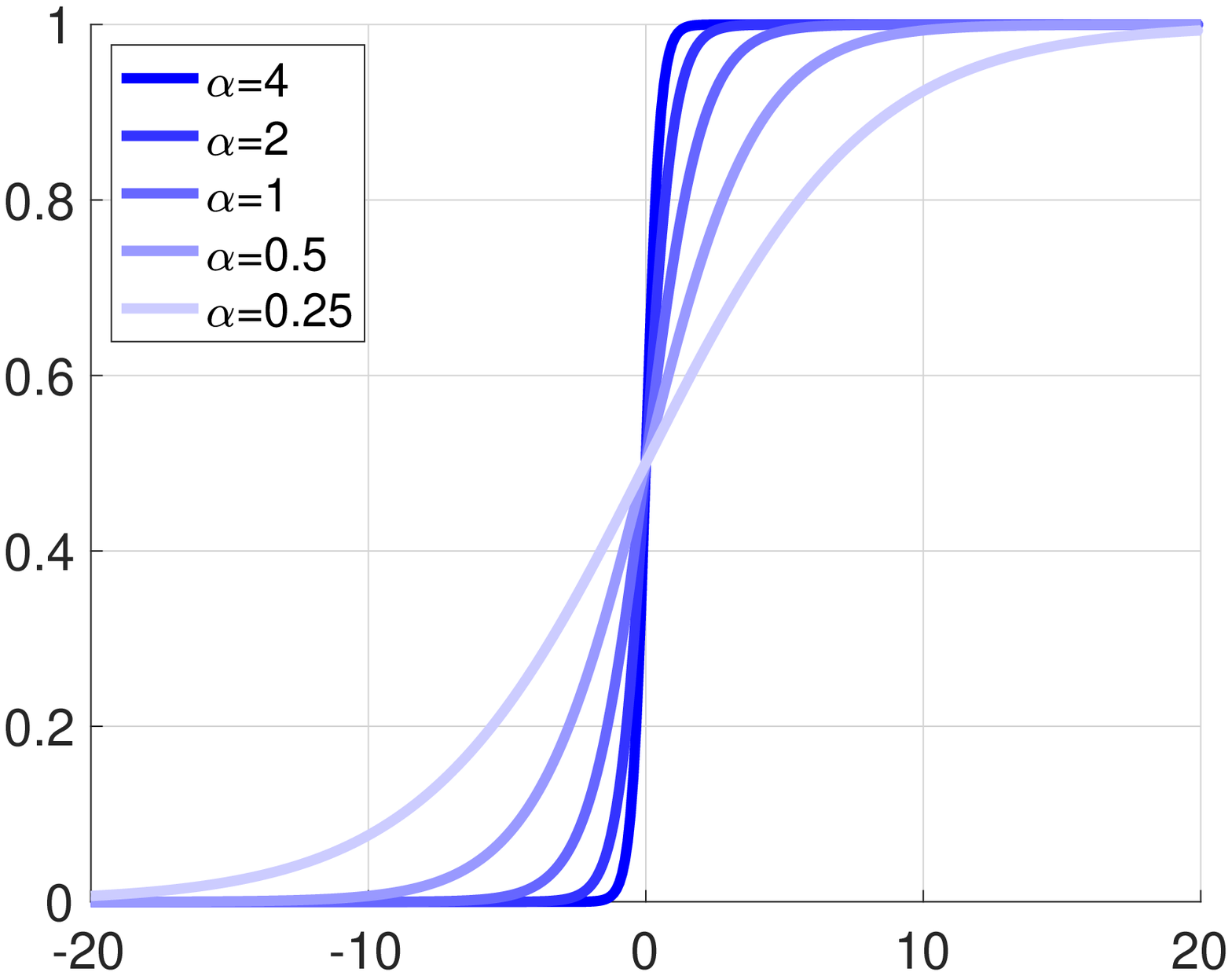}\\
(a) varying $s$ (fixed $\alpha$) & (b) varying $\alpha$ (fixed $s$)
\end{tabular}
\caption{Graphical illustration of sigmoid function with varying (a) scale parameter $s$ with fixed $\alpha$ and (b) steepness parameter $\alpha$ with fixed $s$.}
\label{fig:scale_curve}
\end{figure}
The graphical illustration of the sigmoid function with varying parameters is presented in Fig.~\ref{fig:scale_curve} where the functions with varying $s$ and fixed $\alpha = 1$ are shown in (a), and the functions with varying $\alpha$ and fixed $s = 1$ are shown in (b). 
The update of parameters using the stochastic gradient descent based on mini-batch $\beta^t$ at each iteration $t$ reads: 
\begin{linenomath*}
    \begin{align}
    w^{t+1} \coloneqq w^{t} - \eta^{t} \frac{1}{|\beta^t|} \sum_{i \in \beta^t} \nabla g_i(w^t), \label{eq:update:adaptive:smooth}
    \end{align}
\end{linenomath*}
where the computation of stochastic gradient $\nabla g_i(w^t)$ involves the diffusion $u_i(w^t)$ of residual $d_i(w^t)$. The diffusivity of the heat equation applied to the residual is determined based on a distribution formed by the sigmoid function and its associated parameters, scale $s$ and steepness $\alpha$, are chosen by global and local properties of residual in the neural network architecture as presented in the following section.
%
%
\subsection{Annealing of Adaptive Diffusion}\label{sec:annealing}
The proposed algorithm aims to impose adaptive regularization depending on the magnitude of residual by applying spatially varying diffusion to the residual. The diffusivity of the heat equation applied to the residual is determined based on the magnitude of residual following the sigmoid function as follows:
\begin{linenomath*}
    \begin{align}
    \kappa_i^t = S(d_i(w^t); s^t, \alpha) = \frac{s^t}{1 + \exp(- \alpha \, d_i(w^t))}, \label{eq:diffusivity}
    \end{align}
\end{linenomath*}
where the diffusivity map $\kappa_i^t \in \R^\dimout$ of the heat equation is determined by the sigmoid function of residual $d_i(w^t) \in \R^\dimout$ at iteration $t$.
We consider the temporal residual $d_i(w^t)$ given by the current state of solution $w^t$ in determining the degree of temporal diffusion $k_i^t$ for each sample $(x_i, y_i)$.
We also consider the scale parameter $s^t$ that is variable in optimization time and present its annealing scheme in the following section.
%
%
\subsubsection{Global Adaptivity}
\begin{figure}[htb]
\centering
\begin{tabular}{c@{\hspace{10pt}}c}
\includegraphics[height=80pt]{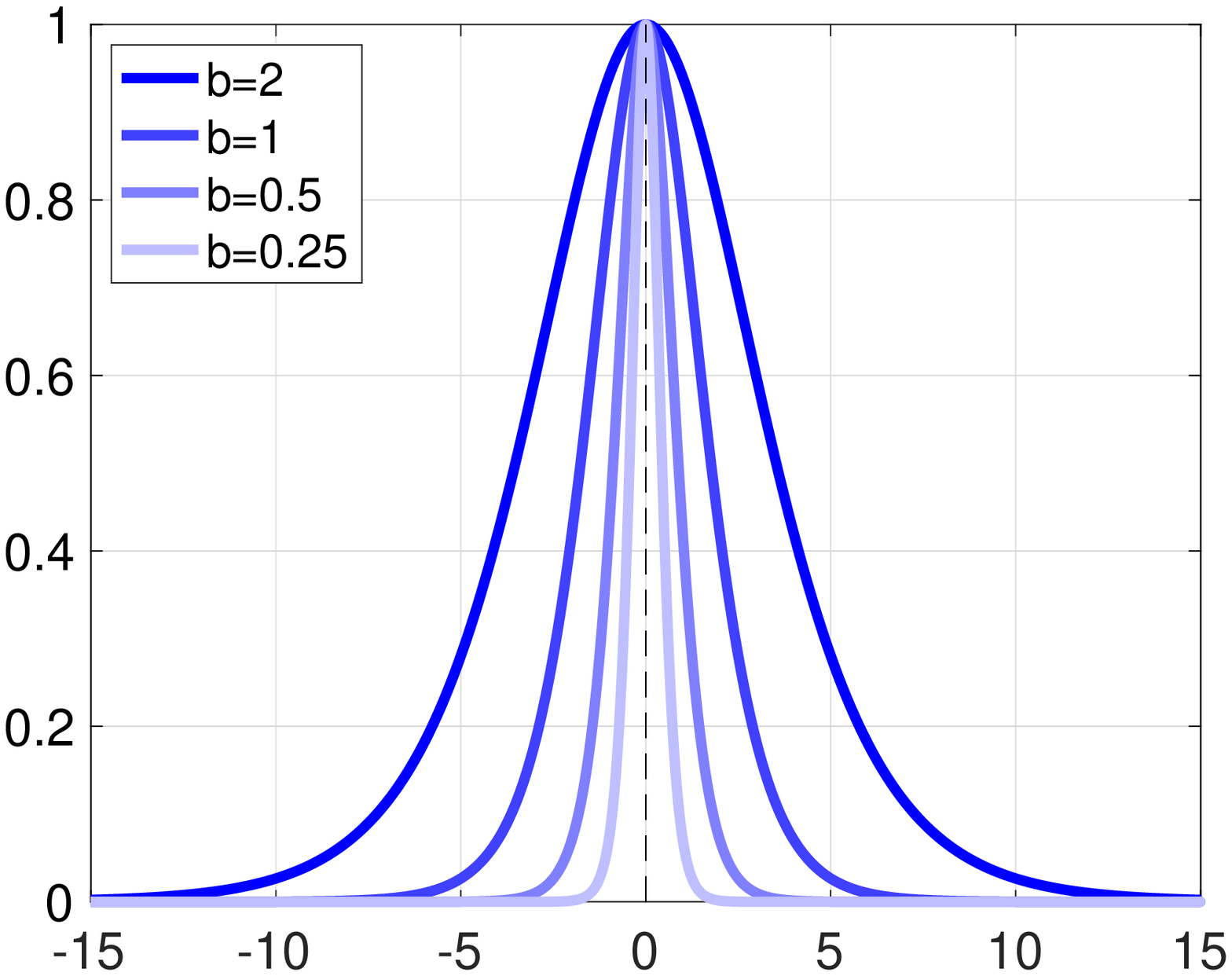} &
\includegraphics[height=80pt]{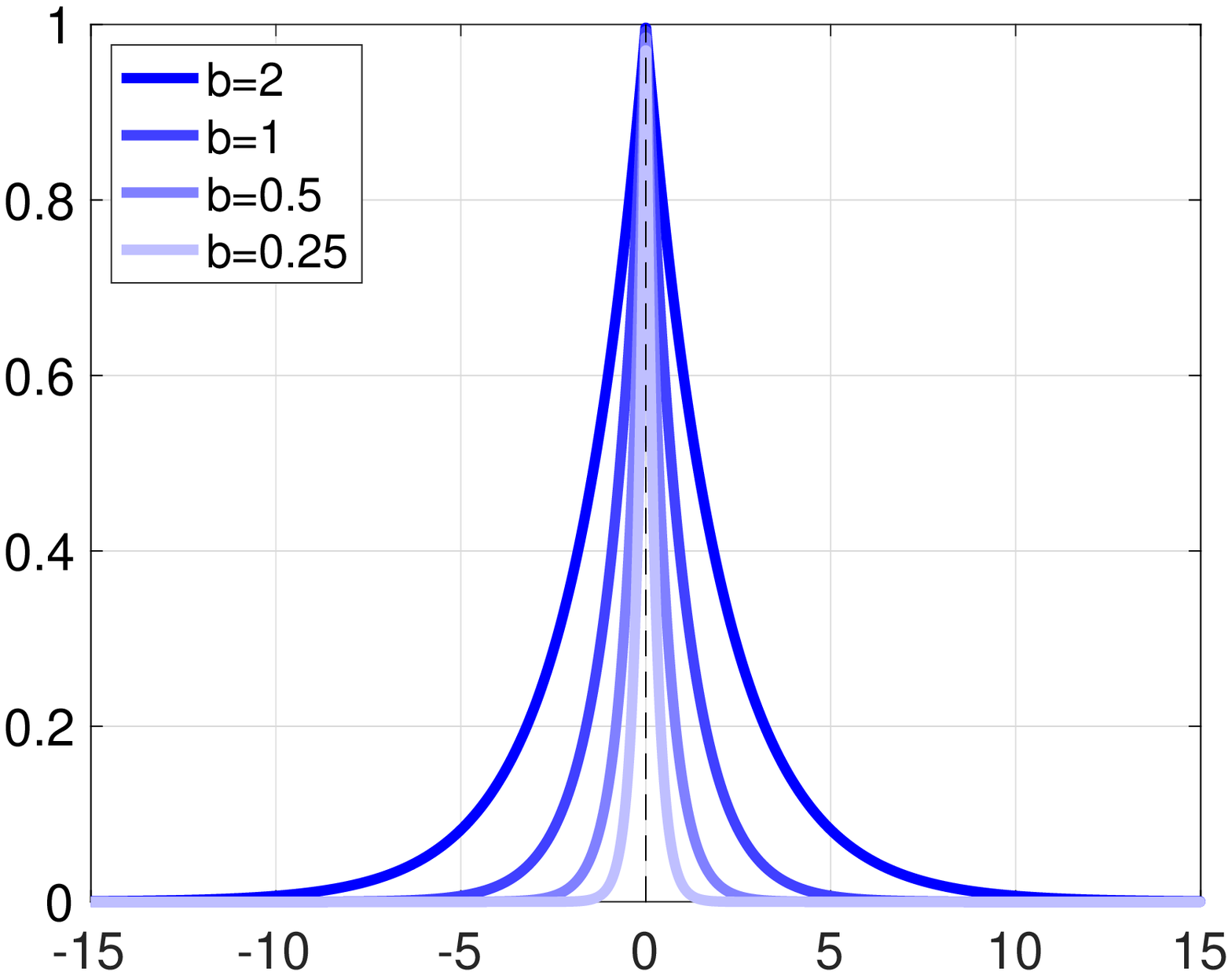}\\
(a) Logistic distribution & (b) Laplace distribution
\end{tabular}
\caption{Graphical illustration of the scaled probability density function associated with different distributions (a) Logistic distribution and (b) Laplace distribution with varying scale $b$ for the annealing of the scale parameter $s$ in sigmoid function. Each probability density function is scaled to have the maximum value 1.} 
\label{fig:annealing:pdf}
\end{figure}
The scale parameter $s^t \in \R^+$ of $S(d_t(w^t); s^t, \alpha)$ at time $t$ in Eq.~\eqref{eq:diffusivity} determines the degree of diffusion that is applied to the entire domain of residual in an isotropic way with fixed steepness parameter $\alpha = 0$, thus it is global parameter that is dependent on time $t$.
The motivation of introducing time-varying scale parameter is to consider the temporal decay of residual resulting in the decrease of diffusion that is equivalent to regularity. 
However, it is often necessary to allow larger stochastic noise in order to avoid undesirable sharp local minima in particular at the early stage of optimization. 
Thus, we propose to employ annealing schemes for the scale parameter $s$ using the probability density functions of either the Logistic distribution $y_1(x ; \mu, b)$ or the Laplace distribution $y_2(x ; \mu, b)$ as defined by:
\begin{linenomath*}
    \begin{align}
    y_1(x; \mu, b) &= \frac{1}{4 b} {\sech}^2\left({\frac{x - \mu}{2 b}}\right), \label{eq:logistic:pdf}\\
    y_2(x; \mu, b) &= \frac{1}{2 b} \exp\left(- \frac{| x - \mu |}{b}\right), \label{eq:laplace:pdf}
    \end{align}
\end{linenomath*}
where $\mu$ and $b$ denote the mean and the scale, respectively. 
The graphical illustrations of the scaled probability density functions $y_1$ and $y_2$ with varying scale parameters $b$ are presented in Fig.~\ref{fig:annealing:pdf} where the maximum value of each probability density function is scaled to have the maximum value 1 and their associated distributions are (a) Logistic and (b) Laplace. 
The global difusivity map $\kappa_i^t$ is then defined by the sigmoid function with $s^t$ and fixed $\alpha = 0$ as defined by:  
\begin{linenomath*}
    \begin{align}
    \kappa_i^t &= S(d_i(w^t); s^t, \alpha), \quad \alpha = 0, \label{eq:diffusivity:global}\\
    s^t &= y(t; \mu, b), \label{eq:scale:global}
    \end{align}
\end{linenomath*}
where $y$ can be either $y_1$ in Eq.~\eqref{eq:logistic:pdf} or $y_2$ in Eq.~\eqref{eq:laplace:pdf}, and $\mu$ is chosen for the peak location and $b$ is a scale parameter for the sharpness of the distribution centered at $\mu$.
The degree of regularization driven by the diffusion process based on the sigmoid function with the annealing for its scale parameter is gradually increasing up to the peak at the mean of the annealing distribution and decreasing afterwards arriving at the original objective function without diffusion. 
%
%
\subsubsection{Local Adaptivity}
In the adaptive application of regularization in the domain of residual, we consider the relative magnitude of residuals so that different degree of regularization is applied to each residual element in its domain.
The residual is initially normalized to have mean $0$ and standard deviation $1$ at each iteration in order to consider the relative significance among the residual elements.
The diffusivity map with the local adaptive scheme is defined by:
\begin{linenomath*}
    \begin{align}
    \kappa_i^t &= S(\tilde{d}_i^t ; s, \alpha), \label{eq:diffusivity:local}\\
    \tilde{d}_i^t &= \frac{d_i(w^t) - \mu_i^t}{\sigma_i^t}, \label{eq:residual:normalize}
    \end{align}
\end{linenomath*}
where parameters $s, \alpha \in \R^+$ are chosen to be constant, and $\tilde{d}_i^t$ is the normalized residual of $d_i^t$ with mean $\mu_i^t$ and standard deviation $\sigma_i^t$ for each sample $(x_i, y_i)$ at time $t$.

%
%
\subsubsection{Combination of Global and Local Adaptivity}
Our final choice of the annealing scheme for adaptive regularization incorporates both global and local approaches considering the global decay of residual and its relative weight in the residual domain at each iteration, leading to the full adaptive scheme. 
The proposed diffusivity map for our algorithm integrates the global annealing of the scale parameter and the relative weight of residual leading to:
\begin{linenomath*}
    \begin{align}
    \kappa_i^t &= S(\tilde{d}_i^t ; s^t, \alpha), \label{eq:diffusivity:full}\\
    \tilde{d}_i^t &= \frac{d_i(w^t) - \mu_i^t}{\sigma_i^t}, \label{eq:residual:full}\\
    s^t &= y(t; \mu, b), \label{eq:scale:full}
    \end{align}
\end{linenomath*}
where $\mu_i^t$ and $\sigma_i^t$ denotes the mean and the standard deviation of temporal residual $d_i(w^t)$ for sample $(x_i, y_i)$, respectively, and $\mu$ and $b$ denotes the mean and the scale for the probability density function of the annealing distribution, respectively.
%
%
%
%
\section{Network Architecture incorporating Adaptive Regularity} \label{sec:netwrok}
\begin{figure}[htb]
\includegraphics[width=\linewidth]{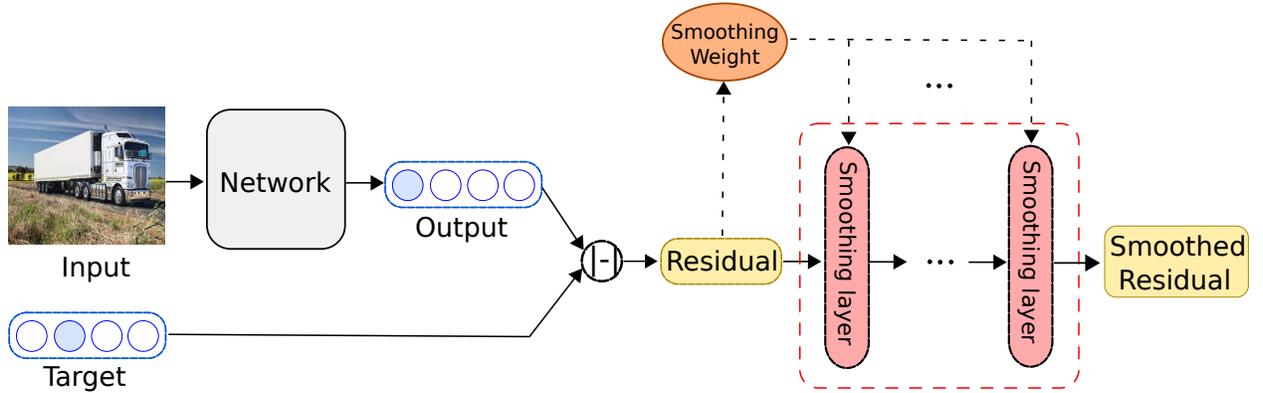}
\caption{Schematic illustration of the network architecture incorporating our regularization algorithm for image classification problem. The target of the primary network is represented by an one-hot encoding and the residual is subsequently fed into a series of smoothing layers leading to the objective function based on the smoothed residual.}
\label{fig:architecture}
\end{figure}
The neural network architecture with our proposed regularization algorithm is constructed by a primary network that yields an output of the prediction for the problem of interest and computes the associated residual that is subsequently fed into a series of smoothing layers leading to the objective function based on the smoothed residual.
The schematic illustration of the network architecture is presented in Fig.~\ref{fig:architecture} where the target of the primary network is represented by a one-hot encoding for the image classification problem. 
Our regularization algorithm applies a diffusion process to the residual depending on its magnitude using the heat equation based on the Laplace operator with spatially varying diffusivity, however the application of the Laplace operator is not suited for the residual domain, in which the spatial property among the neighboring elements is not locally related to the regularity of solution in the image classification problem while the Laplace operator is constrained to be applicable at the residual domain where the local affinity implies the regularity in the solution space, for example, autoencoder architectures.
In the sequel, we employ an extended Laplace operator resulting in a global interpolation of all the elements in the residual domain to blur a one-hot encoding representation based on a fully connected layer with the following weights:
\begin{linenomath*}
\begin{align}
    w_{jk} = 
    \begin{cases}
        1 - \kappa_j, & \text{if } k = j\\
        \frac{\kappa_j}{M-1}, & \text{otherwise }
    \end{cases}
    \label{eq:filter:smoothing}
\end{align}
\end{linenomath*}
where $w_{jk} \in \R^{M \times M}$ denotes the filter element of a fully connected layer that connects from the $k$-th node of the residual to the $j$-th node of the successive diffused residual layer, $M$ is the dimension of residual, and $\kappa_j$ denotes the diffusivity value obtained by Eq.~\eqref{eq:diffusivity:full} for the $j$-th element of the residual.
The number of smoothing layers is related to the diffusion time $\tau$ and the diffusivity $\kappa$ in the heat equation in Eq.~\eqref{eq:heat}, and we set the number of smoothing layer to be one while the scale factor of the diffusivity varies for numerical stability and computational efficiency. The overall algorithm of our proposed method is presented in Algorithm~\ref{algo:smoothing}.

%
%
\section{Experimental Results} \label{sec:exp}
\begin{algorithm*}[htb]
\begin{algorithmic}[1]
\State Initialize the weights $w$ of neural network $h_w$
\State Select annealing scheme $y$
\State Iteration $t=0$
\For {$t=0,1,2,\cdots$}
\State Sample a mini-batch $\beta^t$ from training dataset.
\State Initialize loss $\mathcal{L}(w^t)=0 $
\For{a $i$-th pair of data $(x_i, y_i)$ in $\beta^t$}
\State Get residual $d_i^t= |h_{w^t}(x_i)-y_i|$ and normalize it to get $\tilde{d}_i^t$.  
\State Get diffusivity $\kappa^t_i=S(\tilde{d}^t_i; s^t, \alpha)$ where $s^t=y(t;\mu,b)$
\State Construct a fully connected layer $L$ with the weights $w_{jk}$ as in~\eqref{eq:filter:smoothing}
\State Pass $d_i^t$ to the above fully connected layer $L$ and get the smoothed residual $r_i^t$
\State $\mathcal{L}(w^t)\gets \mathcal{L}(w^t)+\lVert r_i^t\rVert_2^2$ \EndFor
\State $w^{t+1} \gets w^t - \eta^t\frac{1}{|\beta^t|}\nabla \mathcal{L}(w^t)$ where $\eta^t$ is a learning rate at $t$
\State $t\gets t+1$
\EndFor
\State \Return Trained neural network $h_w$
\end{algorithmic}
\caption{Pseudocode for the proposed method}
\label{algo:smoothing}
\end{algorithm*}

\begin{table*}[htb]
	\centering
	\setlength{\tabcolsep}{6pt}
	\begin{tabular}{|c|c|c|c|c|c|c|c|c|}
		\cline{2-9}
		\multicolumn{1}{c|}{}& \multicolumn{4}{c|}{SGD} & \multicolumn{2}{c|}{Global} & \multicolumn{2}{c|}{Global+Local} \\ \hline
		\multicolumn{1}{|c|}{\backslashbox{acc}{wd}}& $1\mathrm{e}{-2}$ & $1\mathrm{e}{-3}$ & $1\mathrm{e}{-4}$ & $1\mathrm{e}{-5}$ &Laplace & Logistic & Laplace & Logistic \\ \hline
		\hline
		max & 88.46 & 93.36 & 93.25 & 93.21 & 93.73 & 93.77 & 93.89 & \textbf{93.91}\\ \hline
		mean & 88.04 & 93.30 & 93.19 & 93.16 & 93.65 & 93.64 & \textbf{93.79} & 93.67\\ \hline
	\end{tabular}
	\caption{Validation accuracy based on ResNet20 using Fashion-MNIST dataset by SGD (left) with varying weight decay (wd) parameters from larger to smaller, our algorithm with global annealing scheme (middle) and the combination of global and local annealing scheme (right). The annealing of adaptive regularization parameter follows Laplace (left) and Logistic (right) distributions where the associated parameters are chosen by the grid search.} 
	\label{tab:fashionMNIST-full}
\end{table*}

\begin{table*}[htb]
	\centering
	\setlength{\tabcolsep}{6pt}
	\begin{tabular}{|c|c|c|c|c|c|c|c|}
		\cline{3-8}
		\multicolumn{2}{c|}{} & \multicolumn{4}{c|}{SGD} &  \multicolumn{2}{c|}{Global+Local} \\ \hline
		 ratio &\multicolumn{1}{c|}{\backslashbox{acc}{wd}}& $1\mathrm{e}{-2}$ & $1\mathrm{e}{-3}$ & $1\mathrm{e}{-4}$ & $1\mathrm{e}{-5}$ &Laplace & Logistic \\ \hline
		\hline
		\multirow{2}{*}{$1/2$} & max & 88.84 & 92.52 & 91.76 & 91.92 & \textbf{92.99} & 92.93 \\ \cline{2-8}
		& mean & 88.55 & 92.45 & 91.73 & 91.87 & \textbf{92.87} & 92.80\\ \hline
		\multirow{2}{*}{$1/4$} & max & 88.46 & 91.32 & 90.55 & 90.58 & \textbf{91.75} & 91.73\\ \cline{2-8}
		& mean & 88.27 & 91.25 & 90.52 & 90.53 & \textbf{91.65} & 91.58\\ \hline
		\multirow{2}{*}{$1/8$} & max & 88.14 & 89.79 & 89.00 & 88.62 & 90.12 & \textbf{90.18}\\ \cline{2-8}
		& mean & 88.02 & 89.72 & 88.94 & 88.55 & 90.01 & \textbf{90.09}\\ \hline
	\end{tabular}
	
	\caption{Validation accuracy based on ResNet20 using partial ($1/2$, $1/4$ and $1/8$) Fashion-MNIST dataset by SGD (left) with varying weight decay (wd) parameters from larger to smaller, our algorithm with the global+local adaptive scheme (right). The annealing of adaptive regularization parameter follows Laplace (left) and Logistic (right) distributions where the associated parameters are chosen by the grid search.} \label{tab:fashionMNIST-partial}
\end{table*}

In the experiments, we empirically provide the quantitative evaluation of our algorithm by the comparative analysis using an image classification task. A detailed description of the experimental setup is presented in the following:

\noindent{\bf Datasets: } We use four commonly used benchmark datasets including CIFAR-10, CIFAR-100~\cite{krizhevsky2009learning}, Street View House Numbers (SVHN)~\cite{netzer2011reading}, and Fashion-MNIST~\cite{xiao2017online}. CIFAR-10 consists of 50K training and 10K testing images of the size 32$\times$32$\times$3 for 10 categories. CIFAR-100 is the same as CIFAR-10 except that it has 100 classes. 
We apply conventional image augmentation with padding, random cropping and flipping to CIFAR-10 and CIFAR-100 as pre-processing stesp.
SVHN is a dataset of house numbers in the street images. It consists of 73257 training and 26032 testing images of the size 32$\times$32$\times$3 for 10 categories. 
Fashion-MNIST consists of 60K training and 10K testing images of the size 28$\times$28 for 10 categories.

\noindent{\bf Neural Network Models: } We consider neural network architectures ranging from shallow to deep models including ResNet20, ResNet56~\cite{he2016deep} and DenseNet-BC with 100 layers ($k=12$)~\cite{huang2017densely}. 

\noindent{\bf Optimization and Hyperparameters: } We use the stochastic gradient descent method and the objective function is the mean squared error of the residual that measures a difference between the prediction and the desired output. 
We use the following common hyperparameters across all the experiments; momentum is 0.9, mini-batch size is 128, number of epoch is 160 for CIFAR-10 and CIFAR-100, 100 for SVHN and 48 for Fashion-MNIST, learning rate is set to be 0.1 for the first 75 percent of epochs and 0.001 for the rest. For Fashion-MNIST dataset, we try several different values ($1\mathrm{e}{-2}, 1\mathrm{e}{-3}, 1\mathrm{e}{-4}, 1\mathrm{e}{-5}$) of weight decay for baseline experiments and the parameter $1\mathrm{e}{-3}$ which shows the best test accuracy among baselines is adopted for the proposed method. For other datasets, weight decay value $1\mathrm{e}{-4}$ is chosen for baseline and proposed method as recommended in~\cite{he2016deep, huang2017densely}.
The unknown weights are initialized by the algorithm proposed in~\cite{he2015delving}.

\noindent{\bf Quantitative Evaluation: } We compute the learning curves that include training loss, training accuracy and validation accuracy. We perform 5 independent trials for each set of experiment and the maximum validation accuracy is taken across all the epochs and the average of the maximum is taken over 5 trials.
We also compute the average validation accuracy over the last 10$\%$ of epochs and the average of the average is taken over 5 trials.

\noindent{\bf Computational cost: } The additional computational cost of our algorithm is $\mathcal{O}(M^2)$ for each training data where $M$ is the dimension of residual. We use a single NVIDA GeForce GTX 1080 Ti GPU. When training CIFAR-10 dataset with ResNet56 architecture, the baseline (SGD) takes about one and half hours whereas our algorithm takes about two and half hours. For CIFAR-10 with DenseNet-BC, baseline takes about four hours whereas our algorithm takes about five hours. Although our algorithm is slower than SGD, it is affordable within a few hours.

\subsection{Ablation Analysis on the Adaptive Regularization Parameters}
\begin{figure*}[htb]
\centering
\begin{tabular}{c@{\hspace{10pt}}c@{\hspace{10pt}}c}
\includegraphics[align=c,height=80pt]{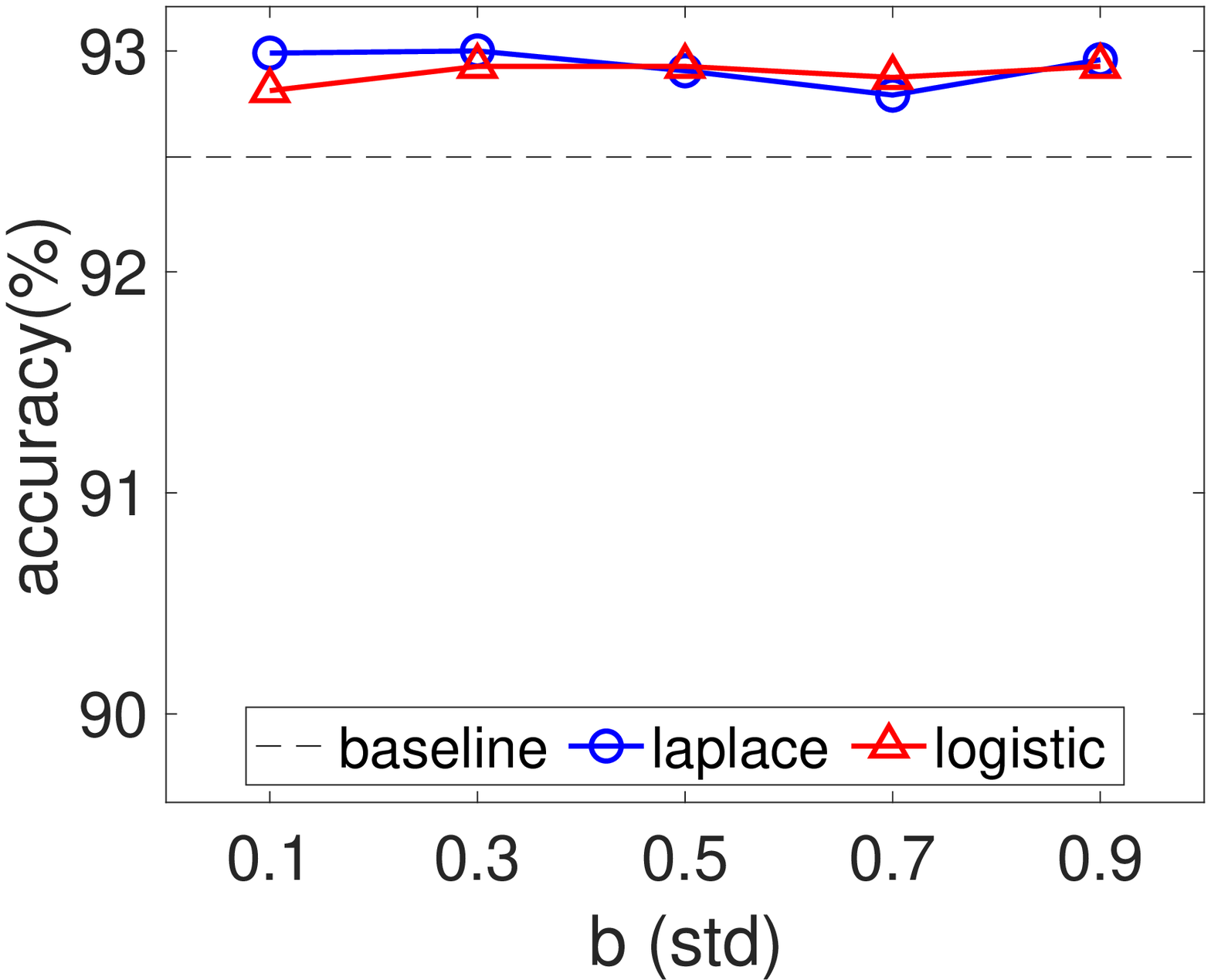} & 
\includegraphics[align=c,height=80pt]{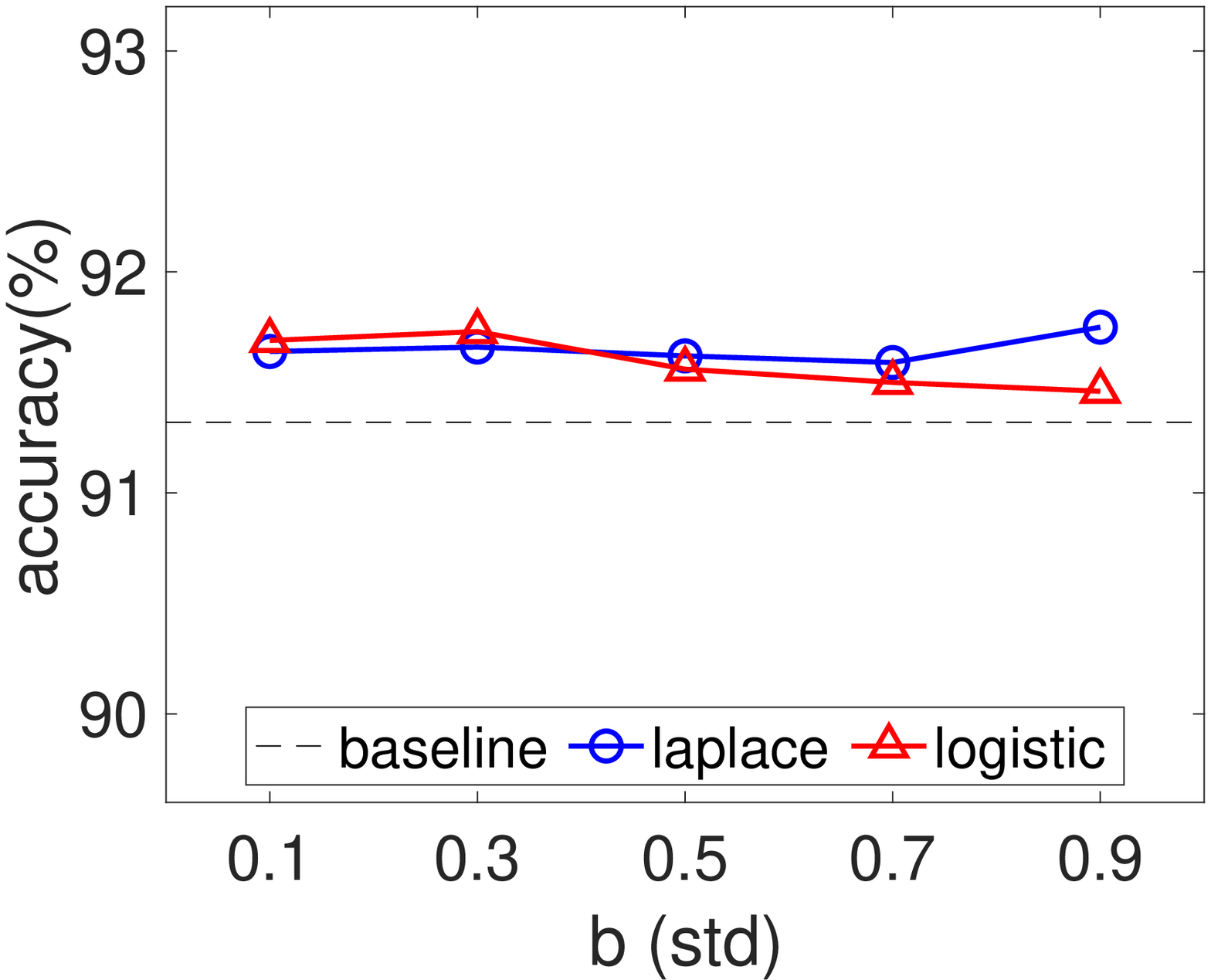} &
\includegraphics[align=c,height=80pt]{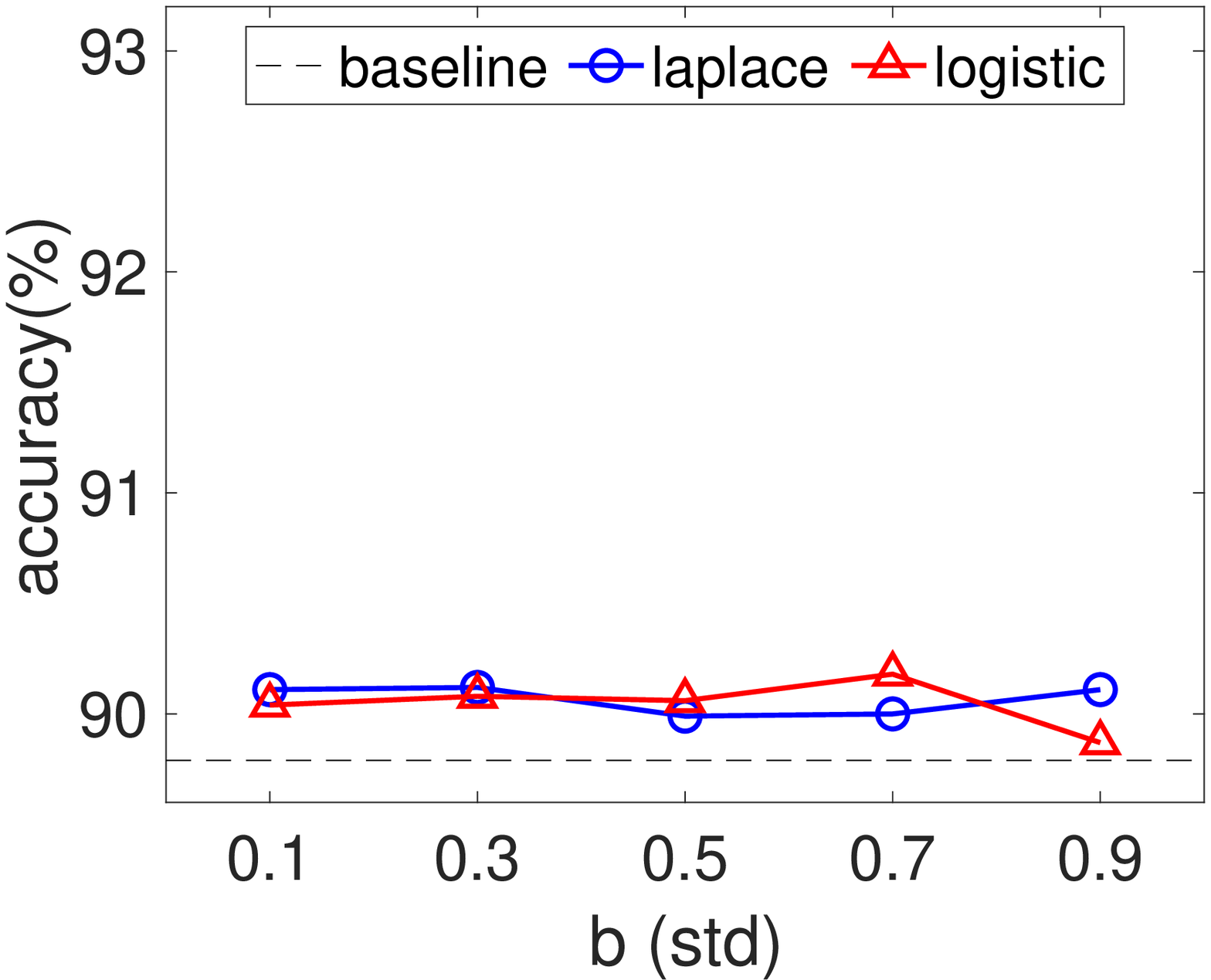} \\

(a) 1/2 training data  & (b) 1/4 training data & (c) 1/8 training data 
\end{tabular}
\caption{Validation accuracy (y-axis) with varying scale parameter of distribution $b$ (x-axis) associated with the annealing distribution using partial Fashion-MNIST dataset. The training is performed based on ResNet20 by SGD and our global+local adaptive regularization schemes. The partial ratios of training data used are (a) $1/2$ (b) $1/4$ and (c) $1/8$.} 
\label{fig:fashionMNIST-partial}
\end{figure*}

\begin{table*}[htb]
	\centering
	\setlength{\tabcolsep}{6pt}
	\begin{tabular}{|c|c|c||c|c|c|c|c|c|}
		\hline
		model & dataset & acc &
		SGD & Adam & AdaGrad & LS~\cite{szegedy2016rethinking} & Laplace & Logistic \\ \hline
		\hline
		\multirow{6}{*}{\rotatebox[origin=c]{90}{ResNet56}} &\multirow{2}{*}{CIFAR-10} & max & 92.66 & 90.50 & 87.67 & 92.36 & \textbf{92.93} & 92.89  \\ \cline{3-9}
		& & mean & 92.54 & 89.86 & 86.47 & 92.21 & \textbf{92.78} & 92.74  \\ \cline{2-9}
		&\multirow{2}{*}{CIFAR-100} & max & 70.94 & 68.42 &  66.55 & 71.09 & 71.06 & \textbf{71.46}  \\ \cline{3-9}
		& & mean & 70.71 & 68.21 & 66.41 & 70.86 & 70.89 & \textbf{71.11}  \\ \cline{2-9}
		& \multirow{2}{*}{SVHN} & max & 95.88 & 95.82 & 94.70 & 95.99 & 96.09 & \textbf{96.11}  \\ \cline{3-9}
		& & mean & 95.83 & 95.37 & 94.36 & 95.95 & 96.05 & \textbf{96.06}  \\ \hline
		\hline
		\multirow{6}{*}{\rotatebox[origin=c]{90}{DenseNet-BC}} &	\multirow{2}{*}{CIFAR-10} & max & 94.43 & 91.91 & 87.16 & 94.43 & 94.64 & \textbf{94.76}  \\ \cline{3-9}
		& & mean & 94.32 & 91.39 & 86.81 & 94.32 & 94.53 & \textbf{94.66}   \\ \cline{2-9}
		&\multirow{2}{*}{CIFAR-100} & max & 75.41 & 67.37 & 58.66 & 75.30 & 75.67 & \textbf{75.75}  \\ \cline{3-9}
		& & mean & 75.22 & 66.75 & 58.44 & 75.12 & \textbf{75.46} & 75.31  \\ \cline{2-9}
		& \multirow{2}{*}{SVHN} & max & 96.82 & 96.11 & 95.10 & 96.86 & 96.89 & \textbf{96.97}  \\ \cline{3-9}
		& & mean & 96.78 & 95.67 & 94.78 & 96.83 & 96.85 & \textbf{96.93}  \\ \hline
	\end{tabular}
	\caption{Comparison of validation accuracy obtained by SGD, Adam, AdaGrad, Label smoothing (LS), our fully adaptive algorithm with Laplace and Logistic distributions from left to right. The training is performed based on the model ResNet56 (top block) and DenseNet-BC (bottom block). For each block of model, the dataset CIFAR-10 (top), CIFAR-100 (middle), SVHN (bottom) are used.}
	\label{tab:state-of-the-art}
\end{table*}

We analyze the effect of global and its combination with local annealing schemes for the adaptive regularization based on ResNet20 using Fashion-MNIST dataset. 
We compare the performance of our algorithm to the baseline, stochastic gradient descent (SGD), to demonstrate that our algorithm outperforms SGD with grid search of regularization parameter.
We apply SGD with varying weight decay values such as $1\mathrm{e}{-2}$, $1\mathrm{e}{-3}$, $1\mathrm{e}{-4}$, $1\mathrm{e}{-5}$ and the validation accuracy is presented in Table~\ref{tab:fashionMNIST-full} where the results with our algorithm based on global adaptive annealing following Laplace and Logistic distributions are presented at the middle block, and the results based on the combination of global and local adaptive annealing following Laplace and Logistic distributions are presented at the right block.
For the global adaptivity, the steepness parameter $\alpha = 0$ in Eq.~\eqref{eq:diffusivity} is used and the scale parameter $b$ of distribution varies from $0.1$ to $0.9$ with step size $0.2$ while the mean $\mu$ of distributions is set to be 75$\%$ point of epochs, the maximum of distributions is scaled to be $1$.
In the application of full adaptive schemes integrating global and local schemes, the same parameters as the global adaptive scheme are used except the steepness parameter $\alpha = 0.25, 0.5, 1, 2, 4$.
We apply a grid search in the selection of parameters associated with our algorithms.
It is shown that our algorithm outperforms SGD regardless of weight decay value associated with SGD, and the performance gain is achieved with the local adaptivity in addition to the global adaptivity.

\subsection{Effect on Generalization based on Partial Training Data}
We empirically demonstrate the effect of our adaptive regularization algorithm based on ResNet20 using Fashion-MNIST dataset. 
We select partial subset of training data uniformly at random for the training phase with varying ratio such as $1/2$, $1/4$ and $1/8$ in highlighting of the effectiveness of our algorithm in generalization.
The validation accuracy of SGD is computed at a range of weight decay values, $1\mathrm{e}{-2}, 1\mathrm{e}{-3}, 1\mathrm{e}{-4}, 1\mathrm{e}{-5}$, and its maximum and average are computed over $5$ independent trials as shown at left block in Table~\ref{tab:fashionMNIST-partial}.
The maximum and average of validation accuracy obtained by our algorithm with fully adaptive regularization incorporating global and local schemes are presented at right block in Table~\ref{tab:fashionMNIST-partial} where Laplace and Logistic distributions are used for global annealing of adaptive regularization.
The associated steepness parameter $\alpha$ with the sigmoid function for the local adaptive regularization is fixed as $1$ whereas the associated scale parameter $b$ with the distribution for the global adaptive regularization is selected by a grid search over a range of values from $0.1$ to $0.9$ with a step size $0.2$ except for Laplace adaptivity scheme with $1/8$ partial data where grid search for $b$ is done over a range from $0.9$ to $1.7$ with a step size $0.2$.

The maximum validation accuracy obtained by our algorithm with different annealing distribution is presented in Figure~\ref{fig:fashionMNIST-partial} where the accuracy with Laplace and Logistic annealing distributions is shown in blue and red, respectively along with the baseline in black for each ratio of partial training set, (a) $1/2$, (b) $1/4$ and (c) $1/8$.  
It is shown that our algorithm outperforms the baseline across all the scale parameters for both annealing distributions, indicating that our algorithm achieves better generalization. 

\subsection{Comparative Analysis with other Optimization Algorithms}

We compare our algorithm with previous smoothing method called label smoothing~\cite{szegedy2016rethinking} and the commonly used optimization algorithms including Adam~\cite{kingma2014adam} and AdaGrad~\cite{duchi2011adaptive}. For label smoothing, we use the same hyperparameters as that of baseline and smoothing parameter $\epsilon=0.1$. 
In our comparative analysis, we use deeper networks including ResNet56 and DenseNet-BC with 100 layers ($k=12$) for the benchmark datasets that are CIFAR-10, CIFAR-100 and SVHN.
The associated parameters with our algorithm are used by $b = 0.2, 0.5$ for Laplace, $b=0.25, 0.5$ for Logistic distribution respectively and $\alpha = 1$.
The maximum and mean validation accuracy are presented in Table~\ref{tab:state-of-the-art} where the results using ResNet56 (top block) and DenseNet-BC (bottom block) with SGD, Adam, AdaGrad, and our algorithms with Laplace and Logistic distributions are shown from left to right.

\begin{figure*}[htb]
\centering
\begin{tabular}{c@{\hspace{5pt}}c@{\hspace{15pt}}c@{\hspace{15pt}}c}
\rotatebox[origin=c]{90}{CIFAR-10} & 
\includegraphics[align=c,height=80pt]{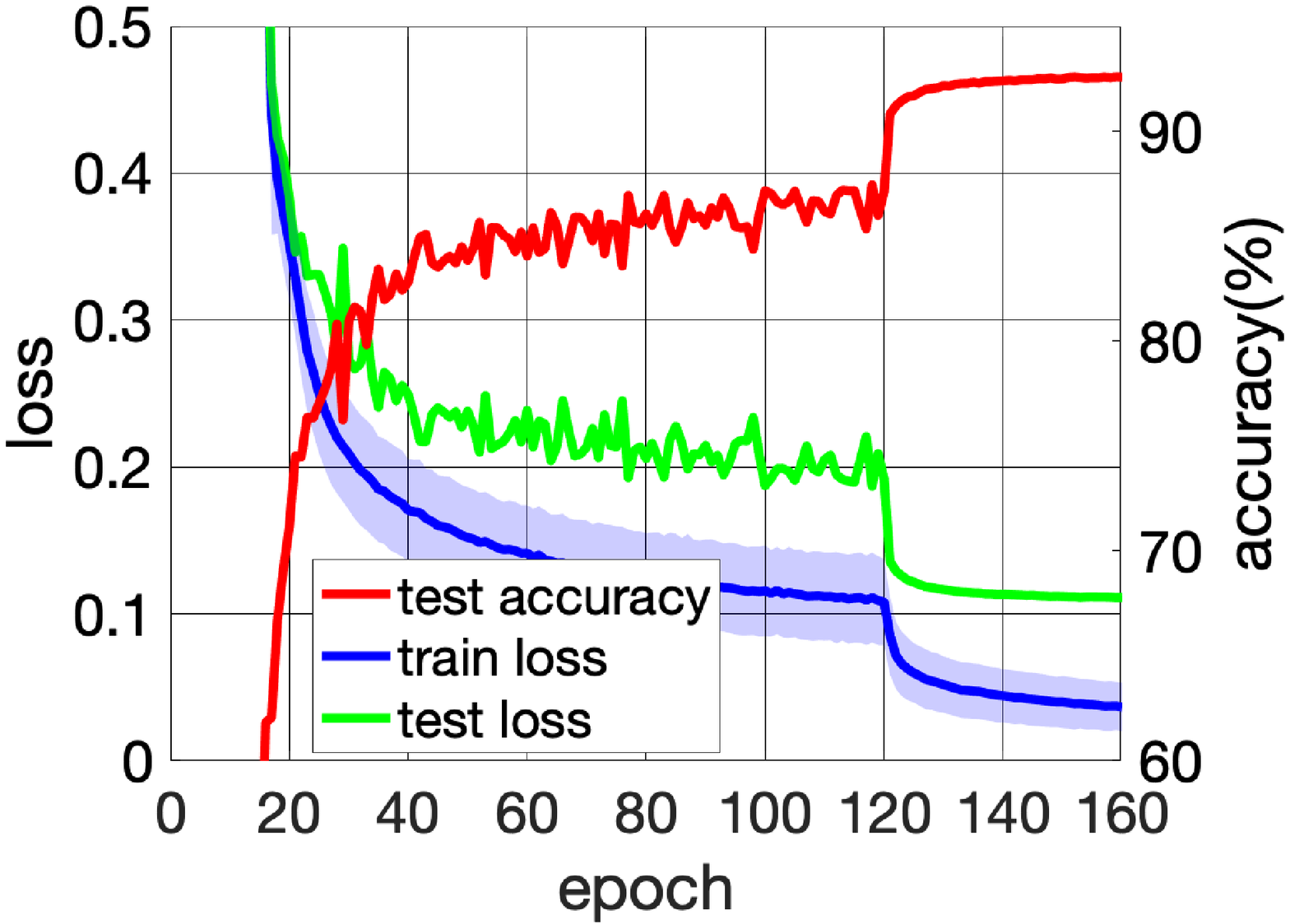} & 
\includegraphics[align=c,height=80pt]{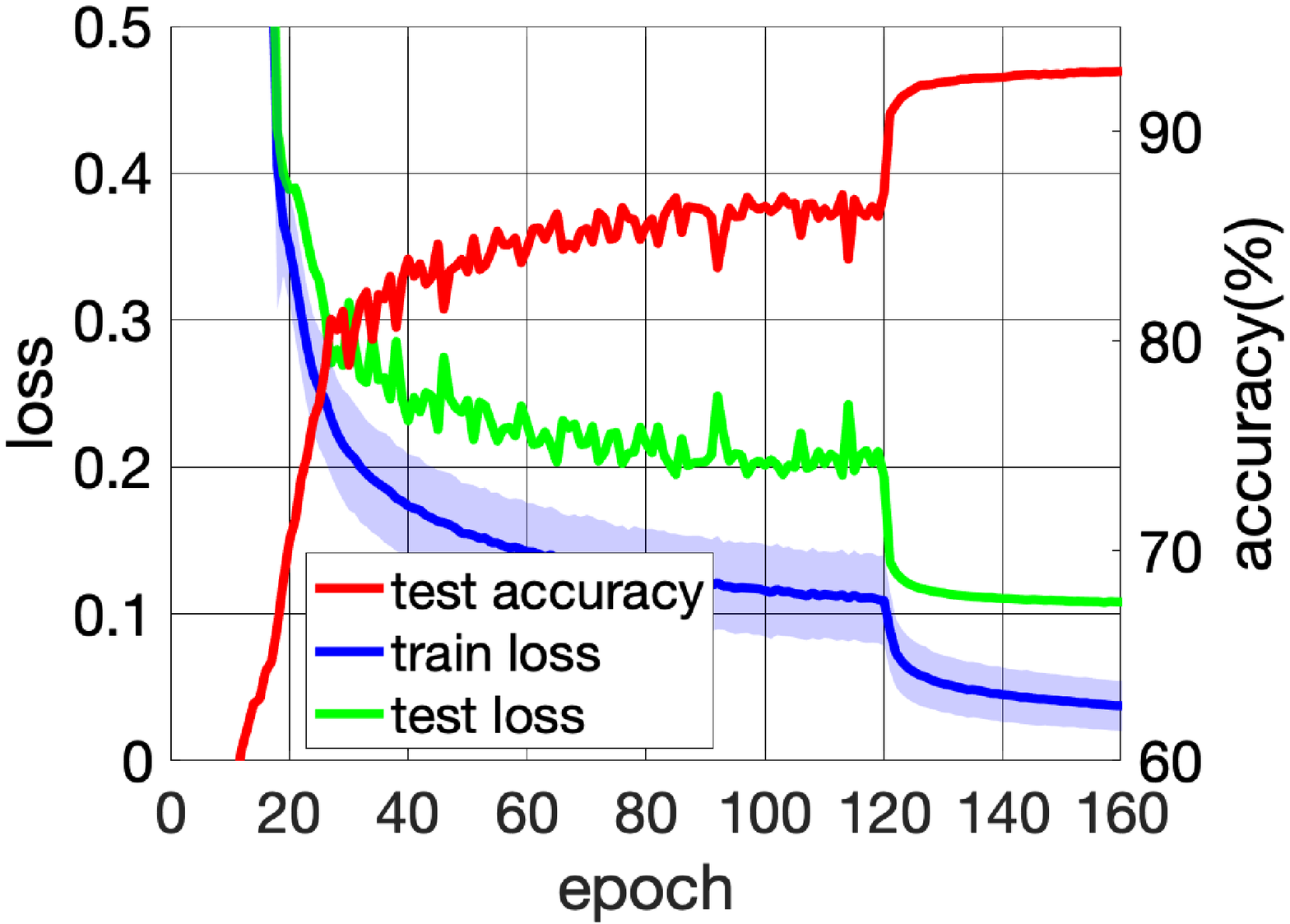} &
\includegraphics[align=c,height=80pt]{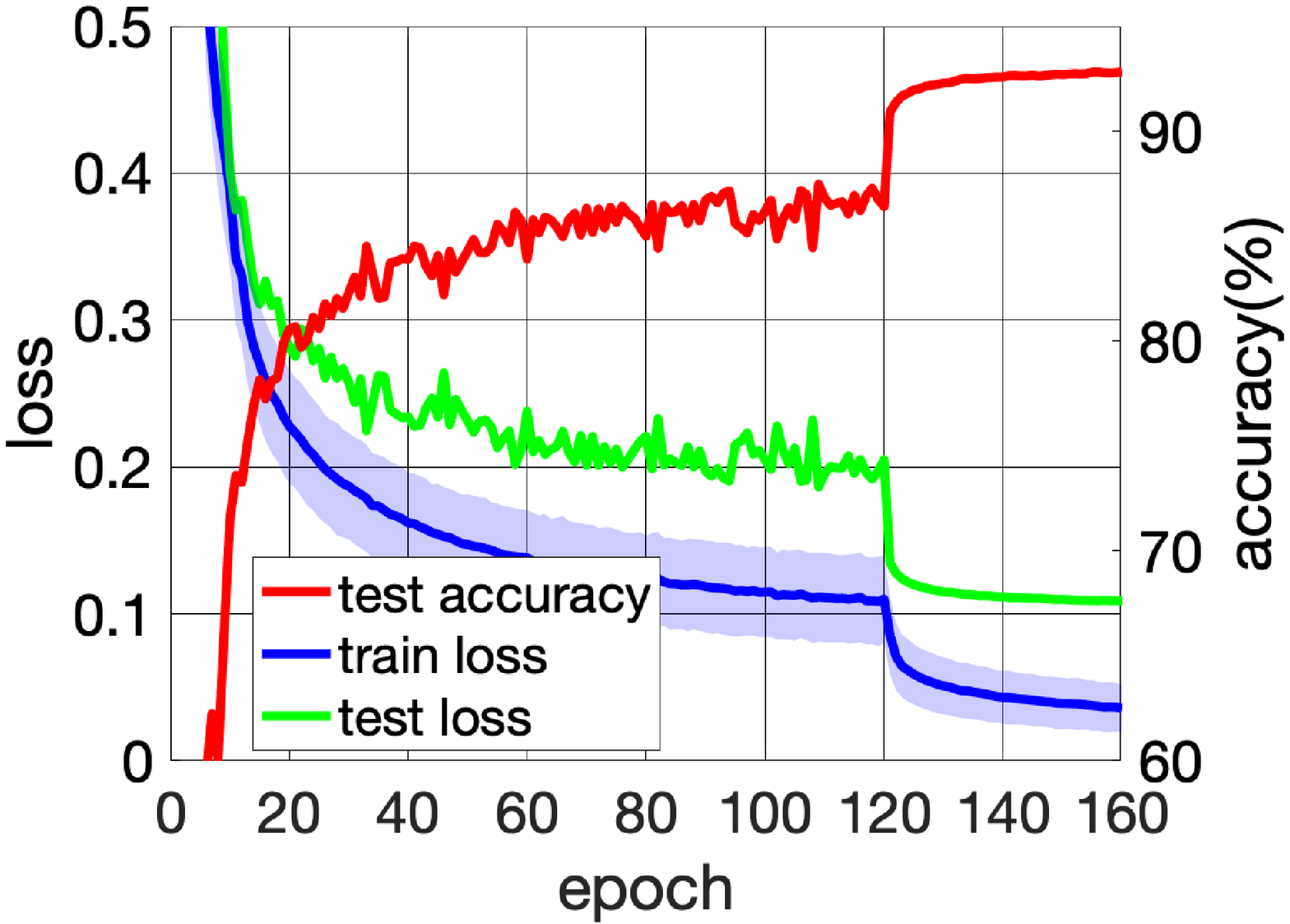}\\

\rotatebox[origin=c]{90}{CIFAR-100} &
\includegraphics[align=c,height=80pt]{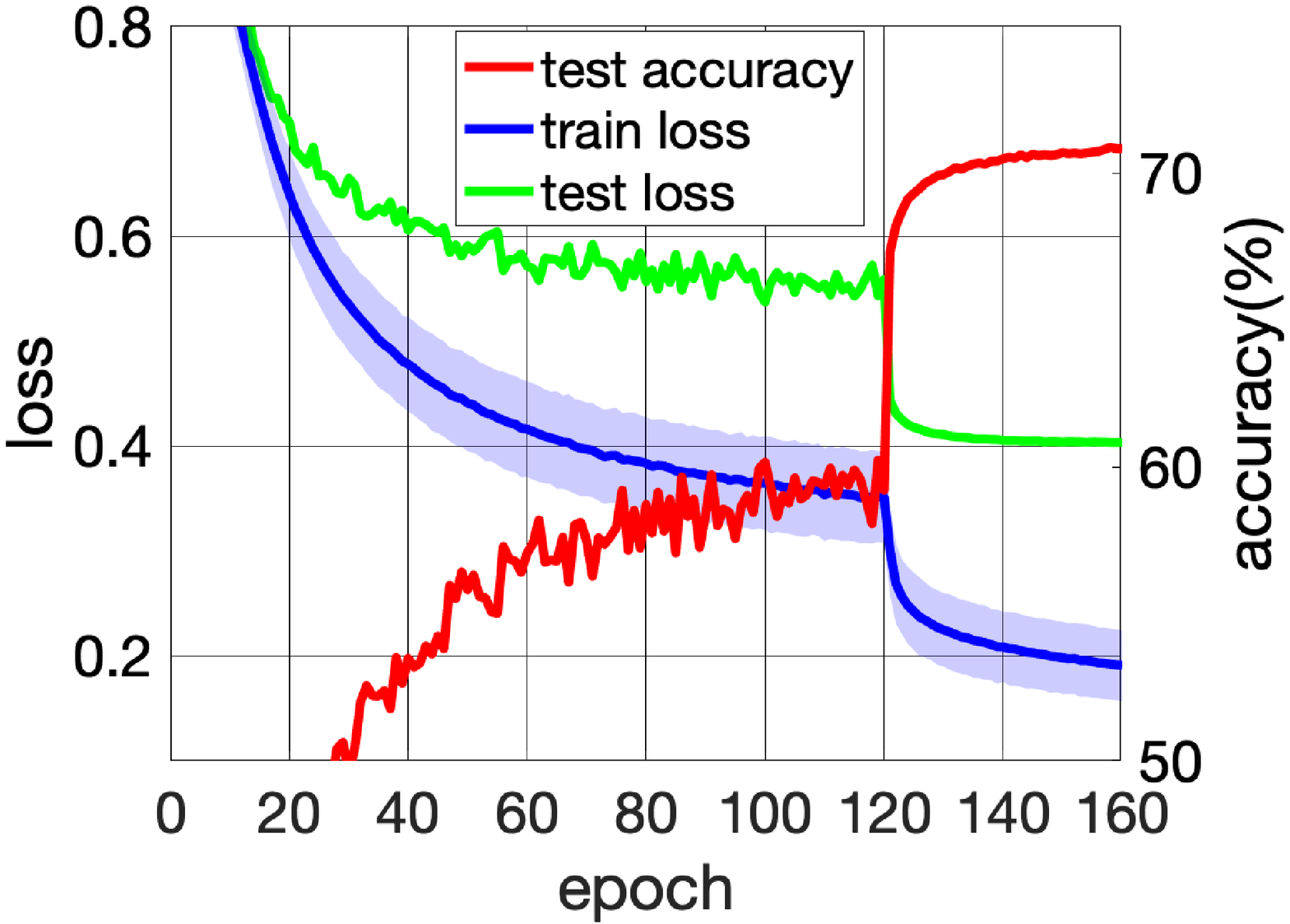} &
\includegraphics[align=c,height=80pt]{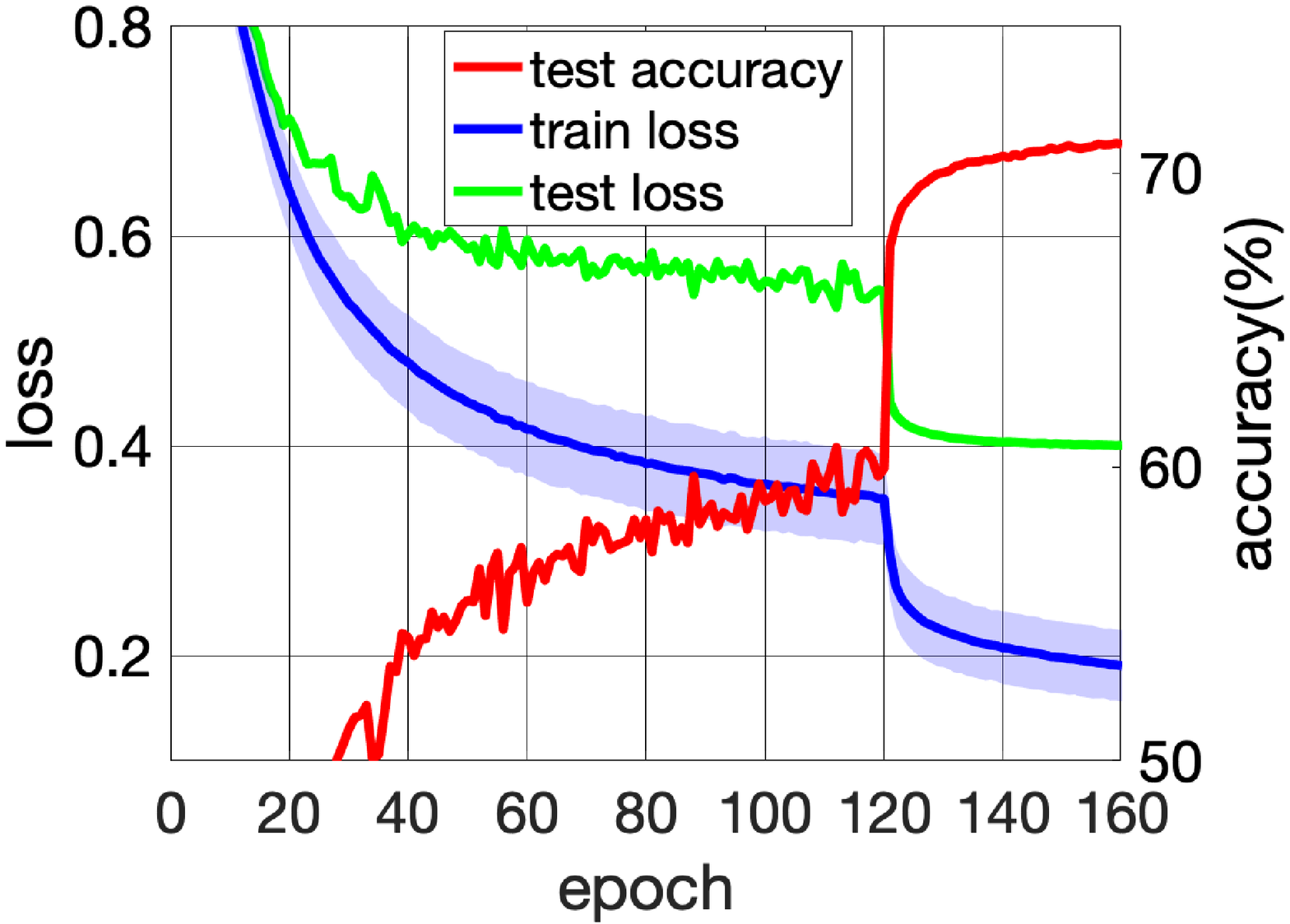} &
\includegraphics[align=c,height=80pt]{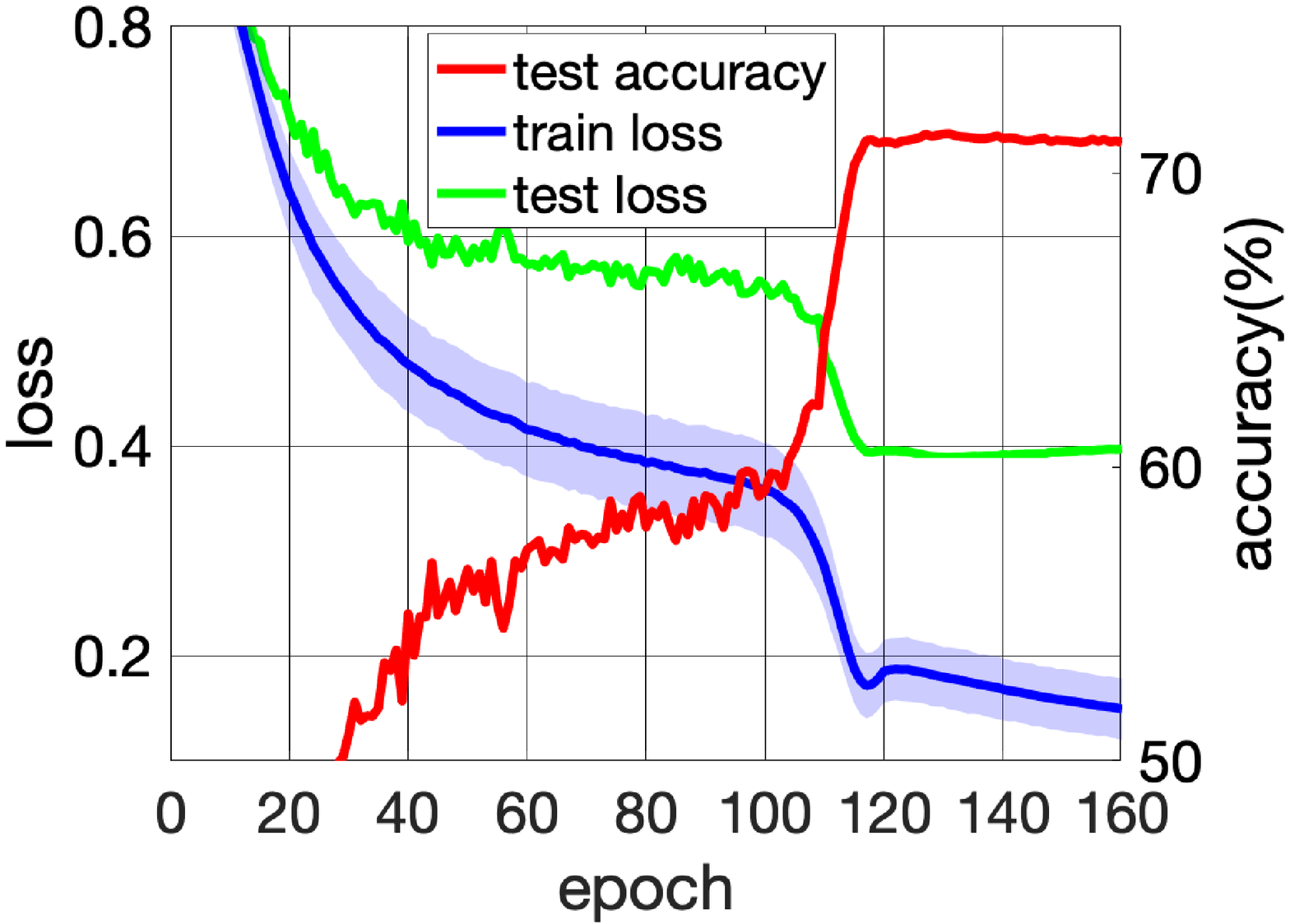}\\

\rotatebox[origin=c]{90}{SVHN} & 
\includegraphics[align=c,height=80pt]{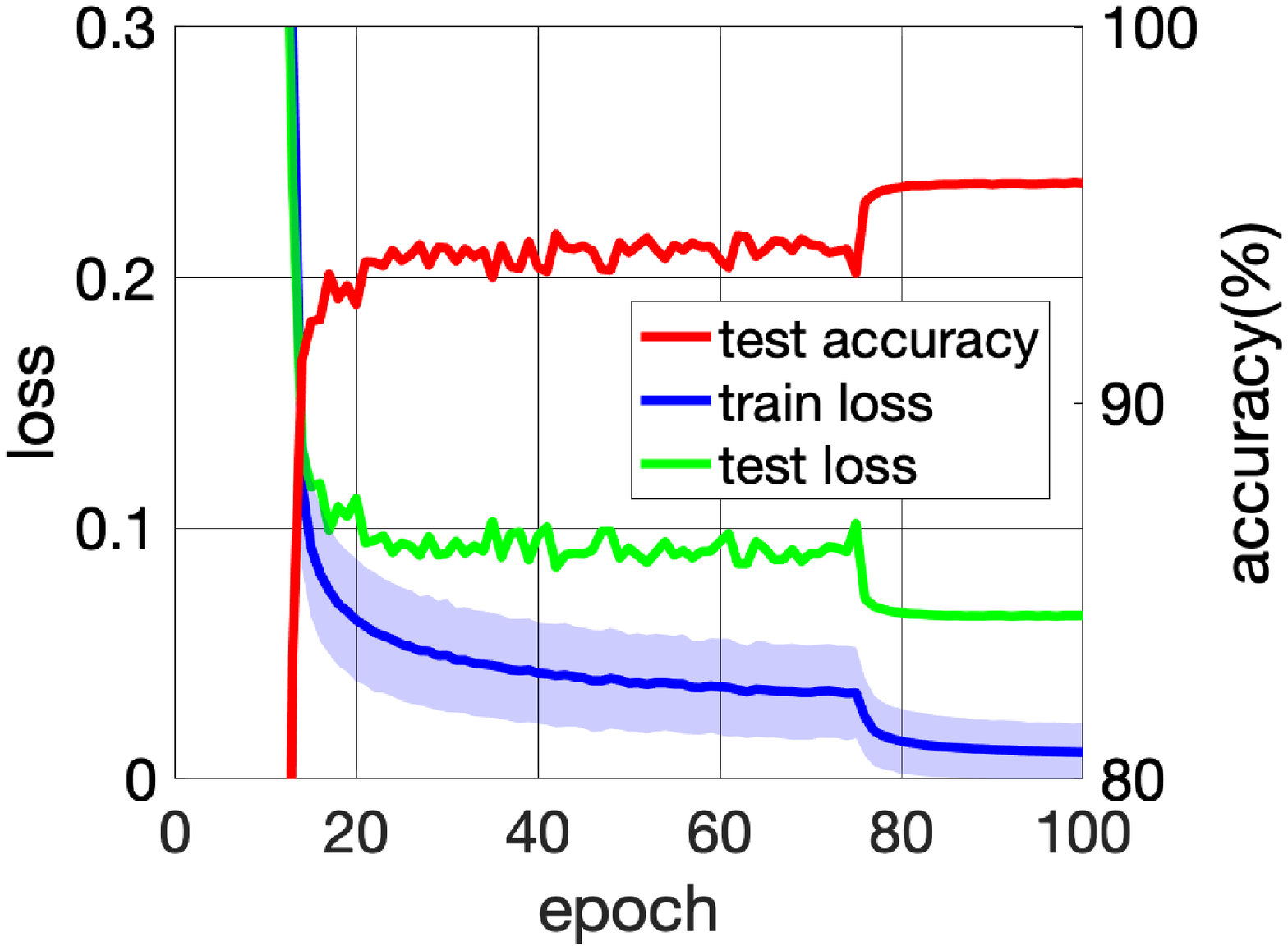} &
\includegraphics[align=c,height=80pt]{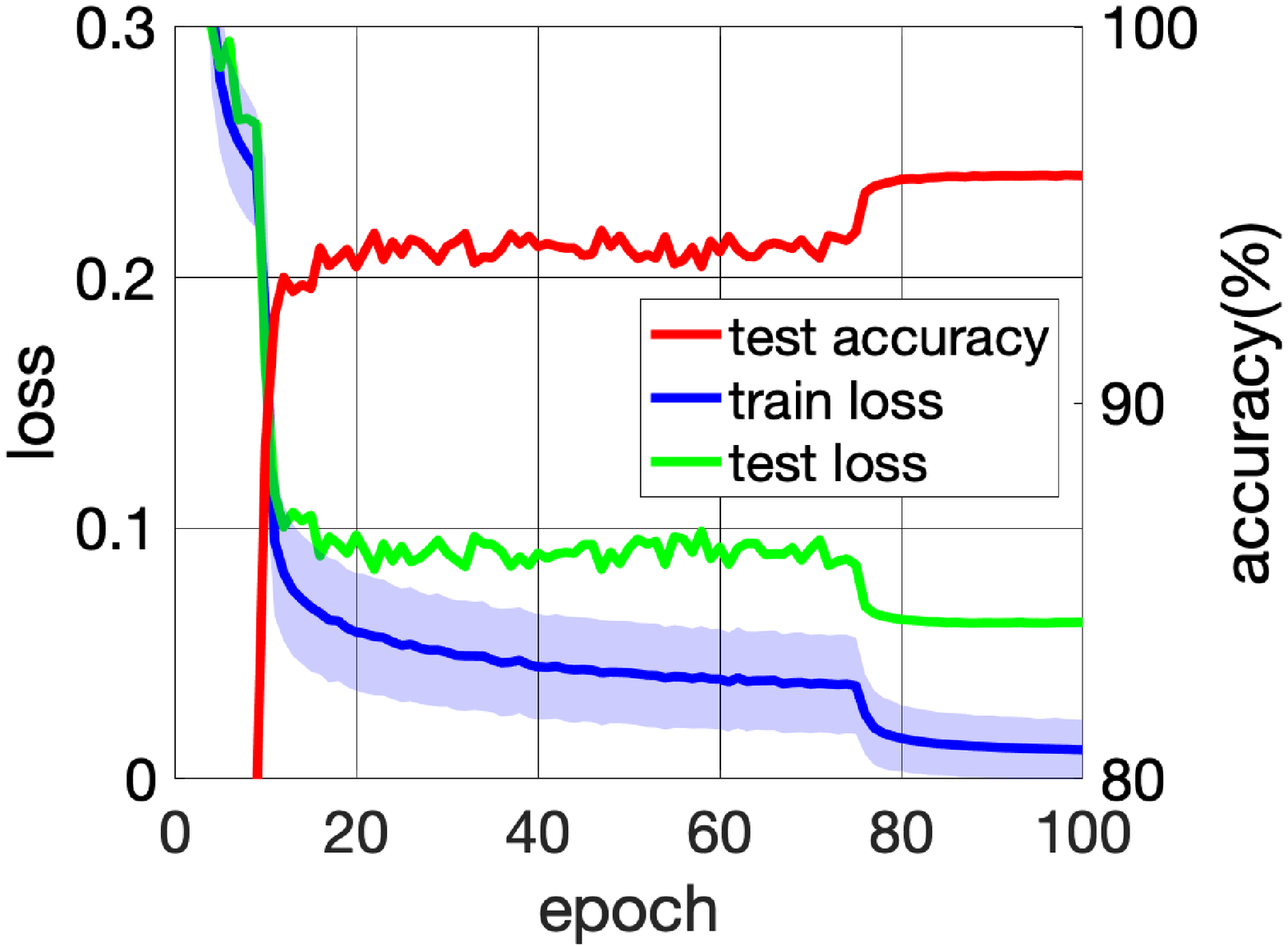} &
\includegraphics[align=c,height=80pt]{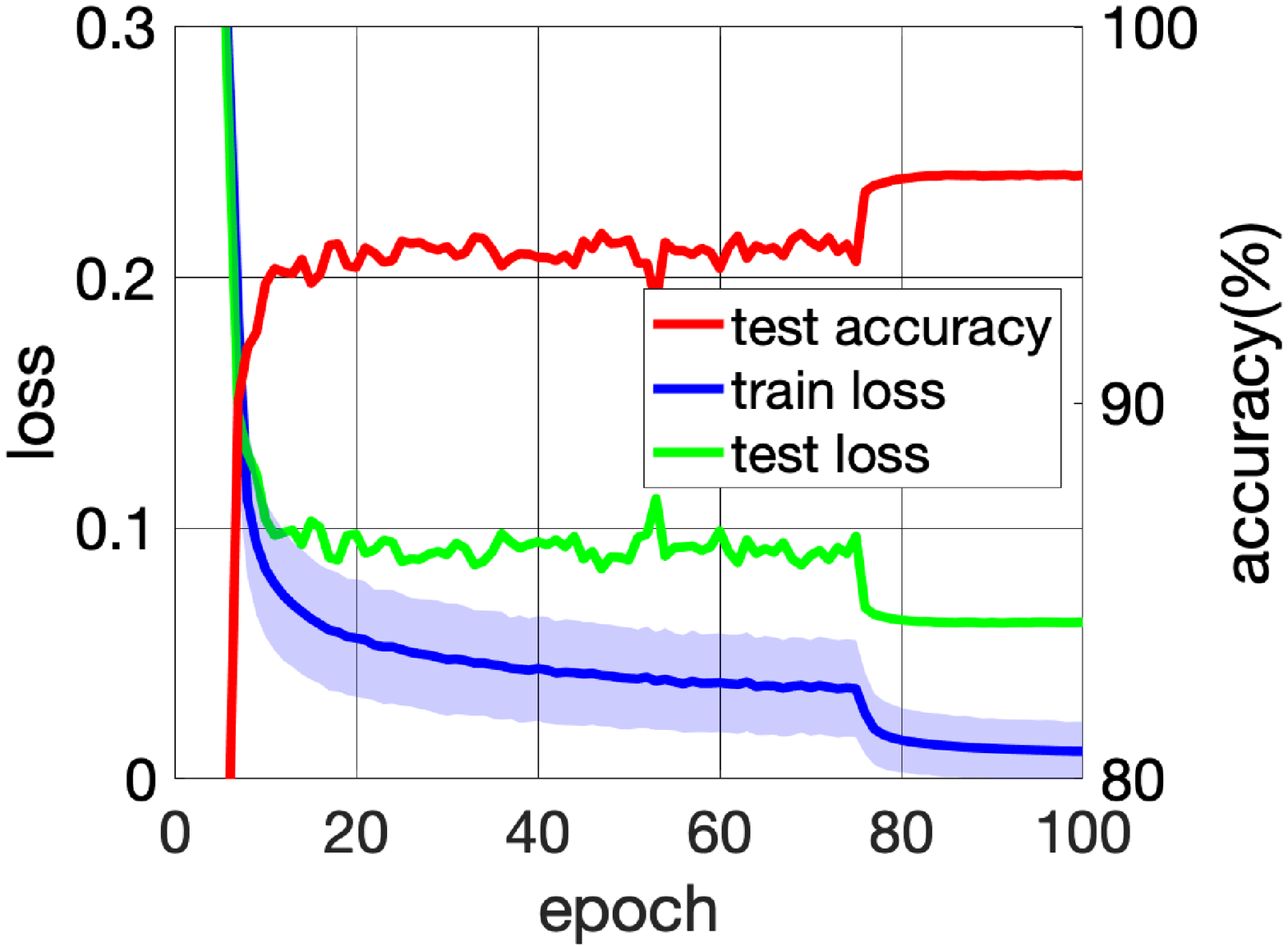}\\
& (a) SGD & (b) Laplace & (c) Logistic
\end{tabular}
\caption{Learning curves obtained based on ResNet56 model using CIFAR-10 (top), CIFAR-100 (middle), and SVHN (bottom) datasets. The validation accuracy, training loss, testing loss are presented in red, blue, green color, respectively. The learning performance of our regularization scheme based on (b) Laplace and (c) Logistic distributions is compared with (a) the SGD algorithm.} 
\label{fig:resnet56}
\end{figure*}

\begin{figure*}[htb]
\centering
\begin{tabular}{c@{\hspace{5pt}}c@{ \hspace{15pt}}c@{\hspace{15pt}}c}
\rotatebox[origin=c]{90}{CIFAR-10} & 
\includegraphics[align=c,height=80pt]{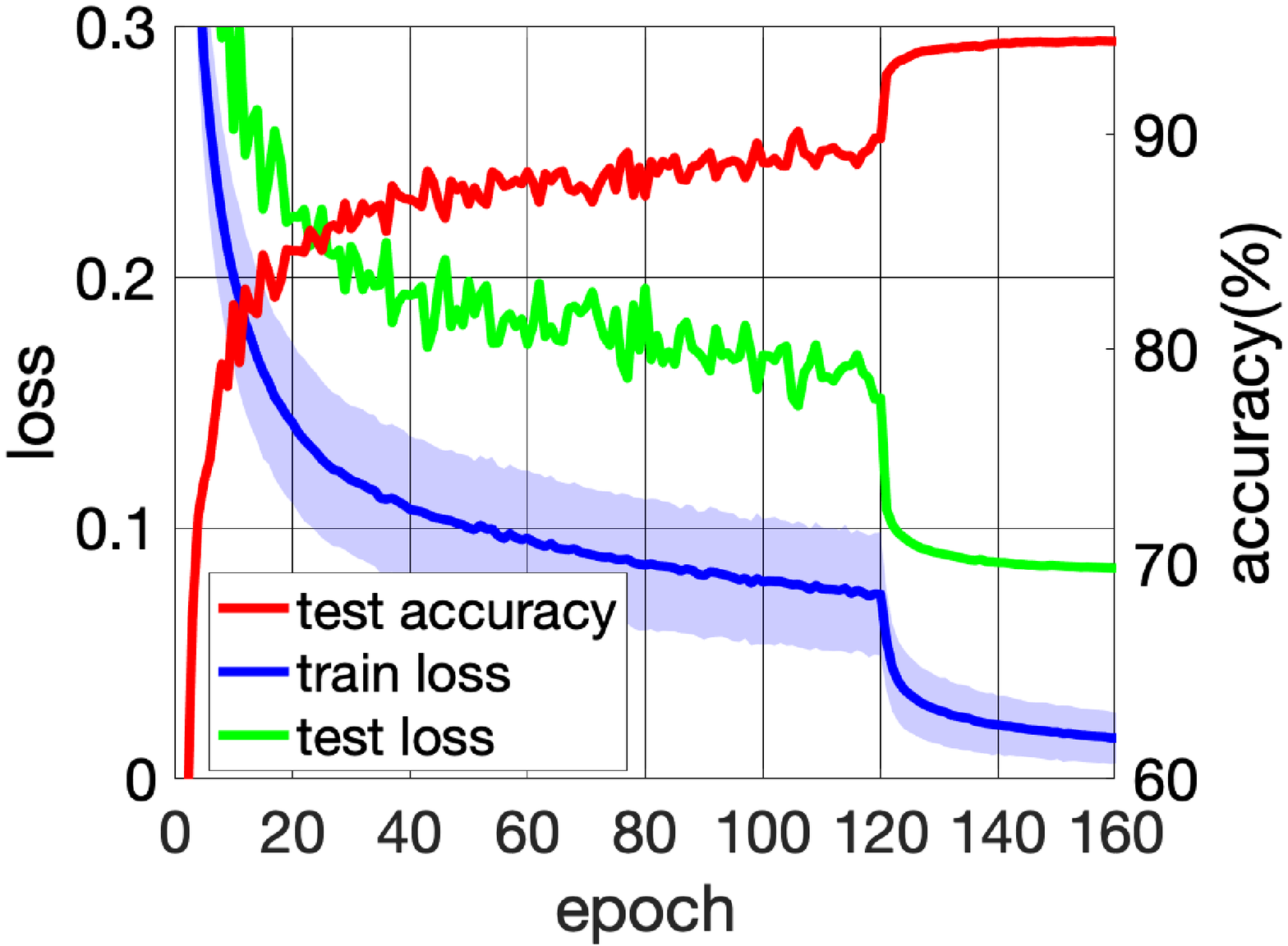} &
\includegraphics[align=c,height=80pt]{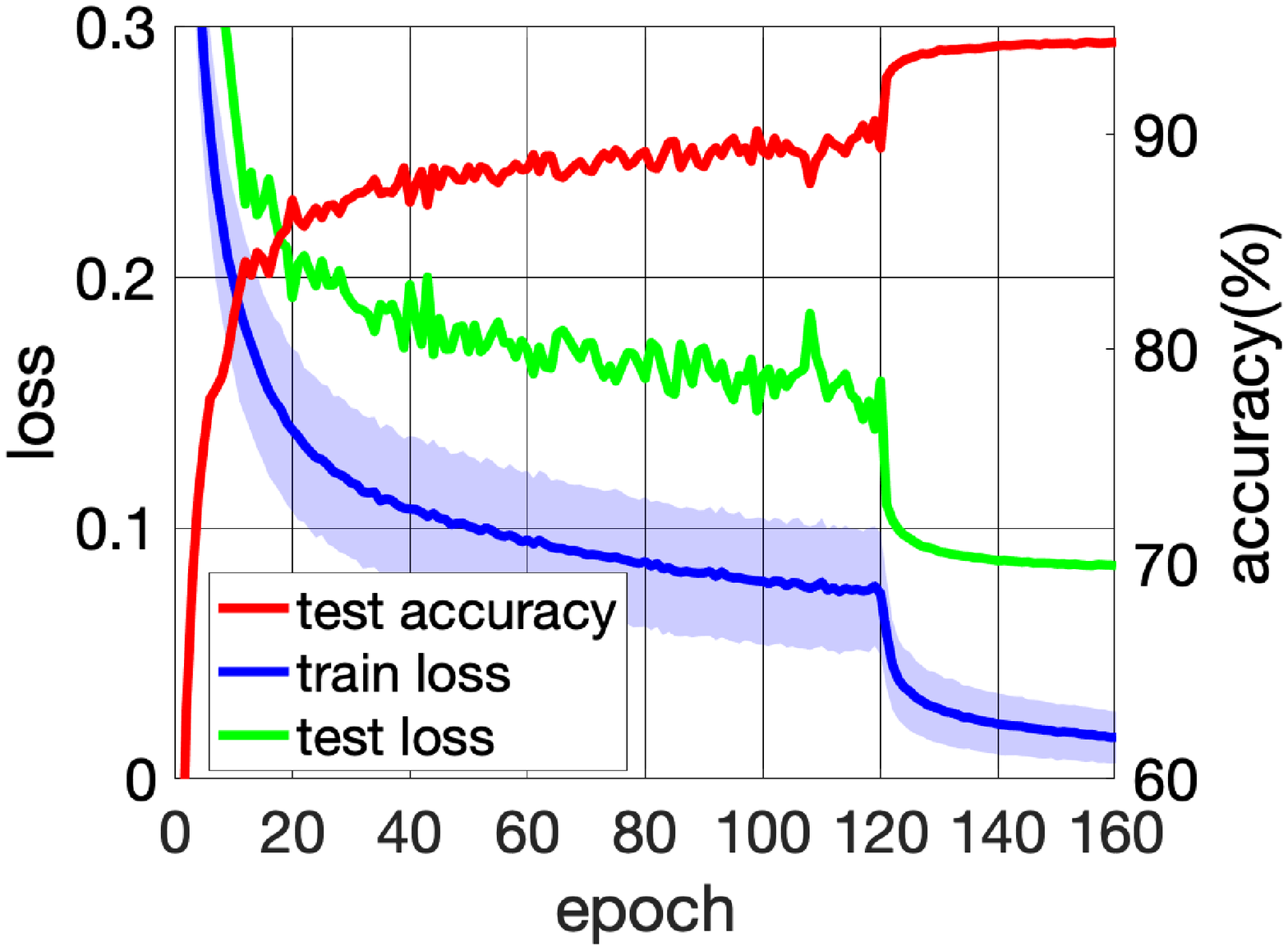} &
\includegraphics[align=c,height=80pt]{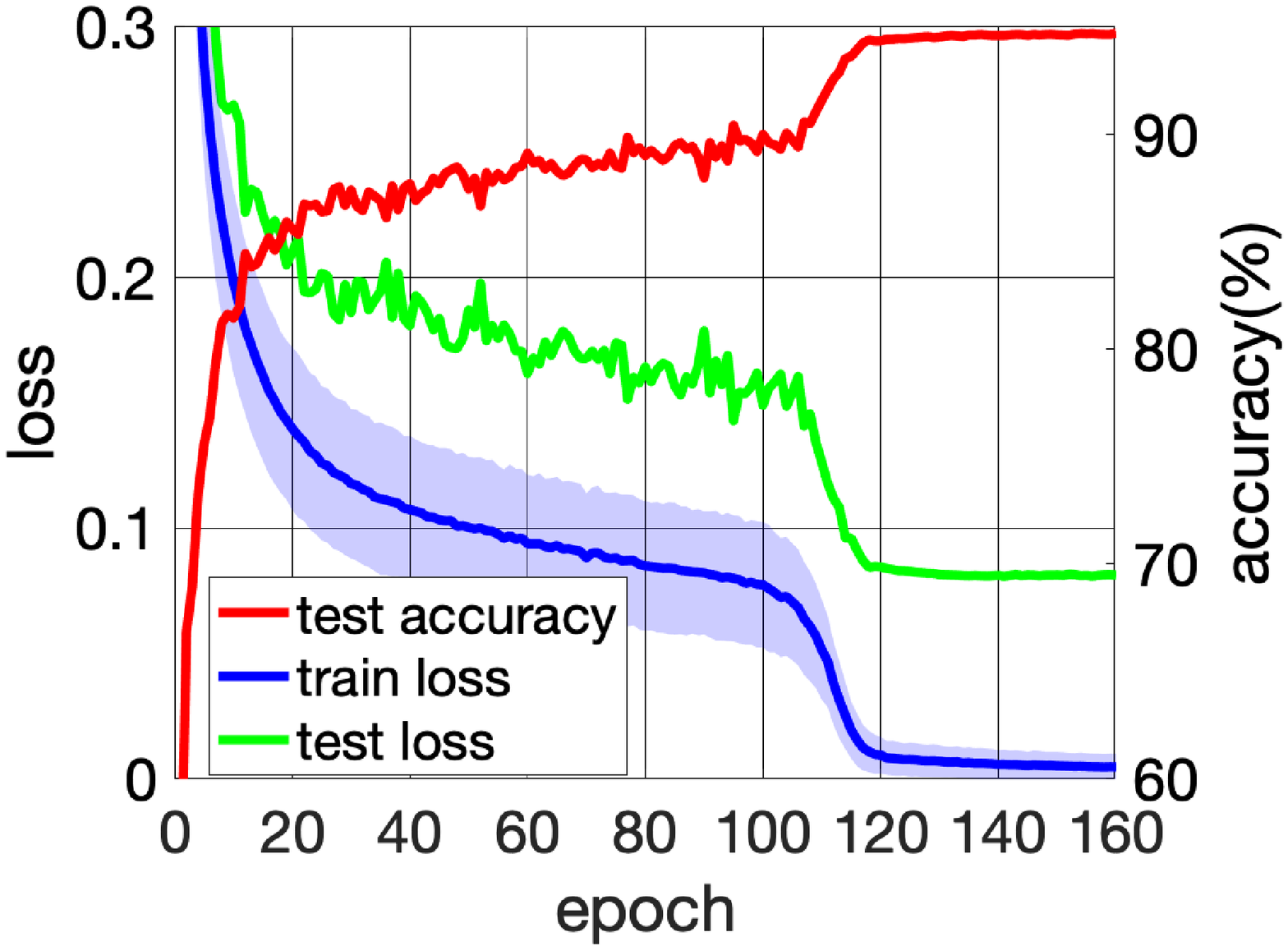}\\

\rotatebox[origin=c]{90}{CIFAR-100} & 
\includegraphics[align=c,height=80pt]{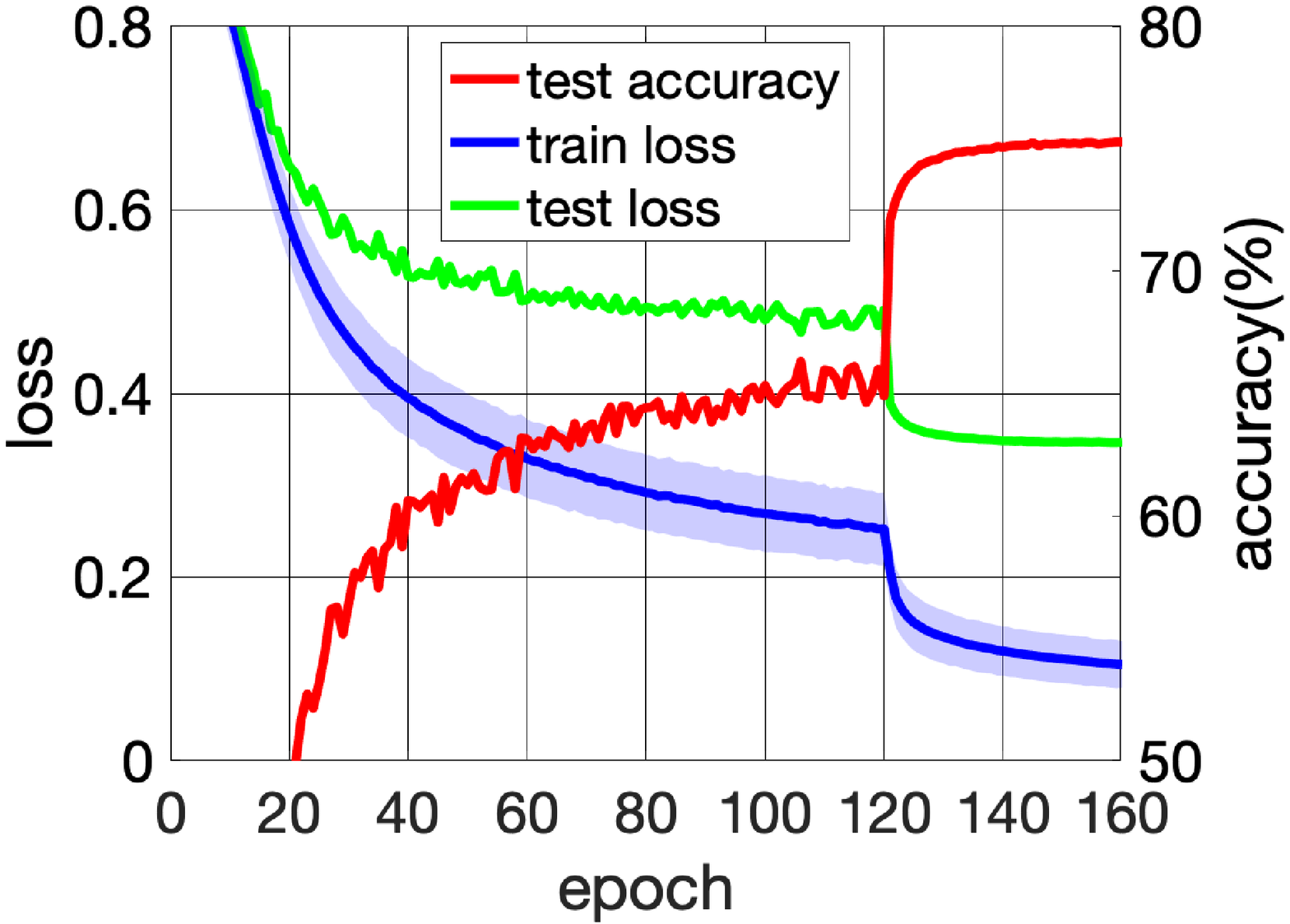} &
\includegraphics[align=c,height=80pt]{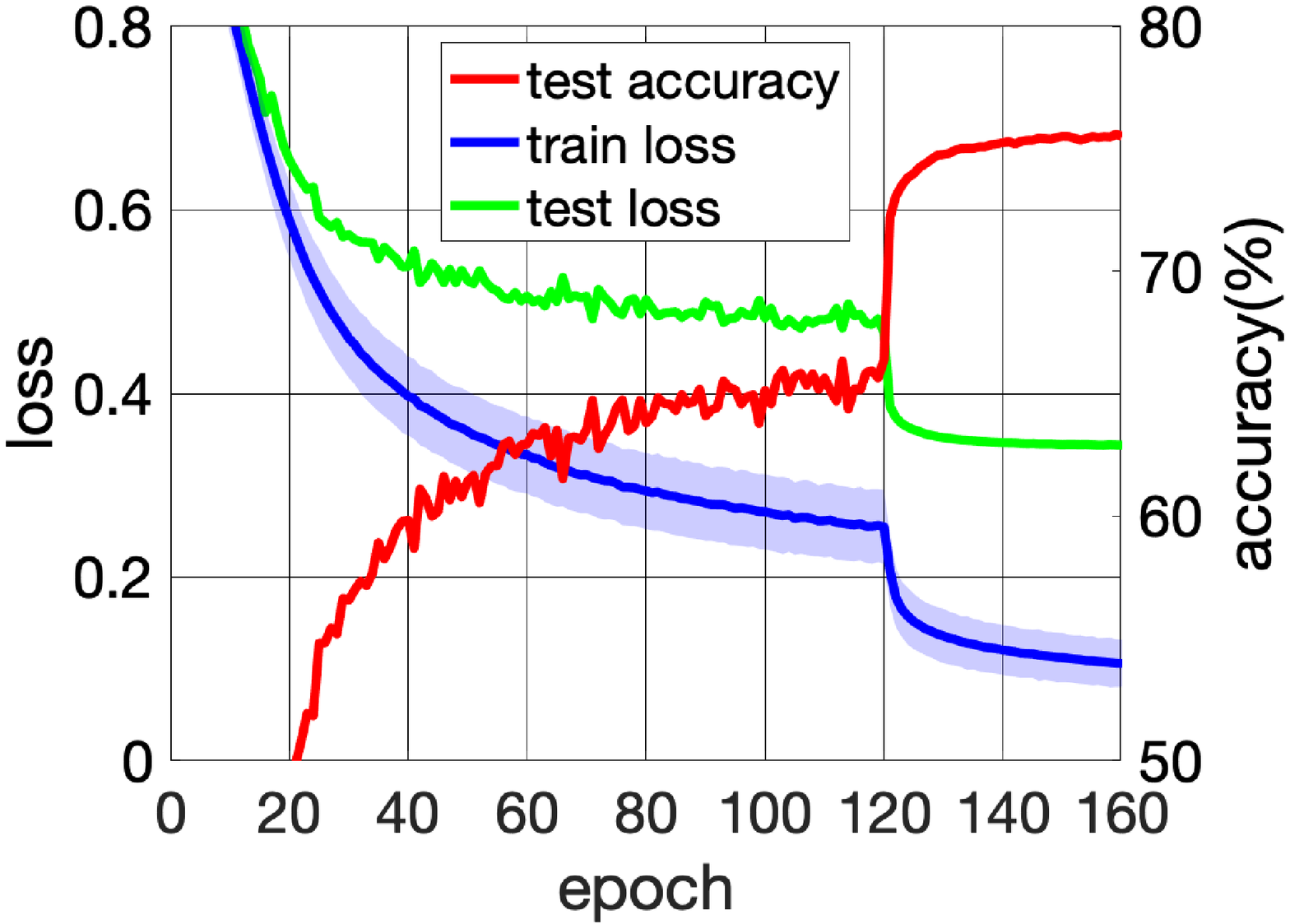} &
\includegraphics[align=c,height=80pt]{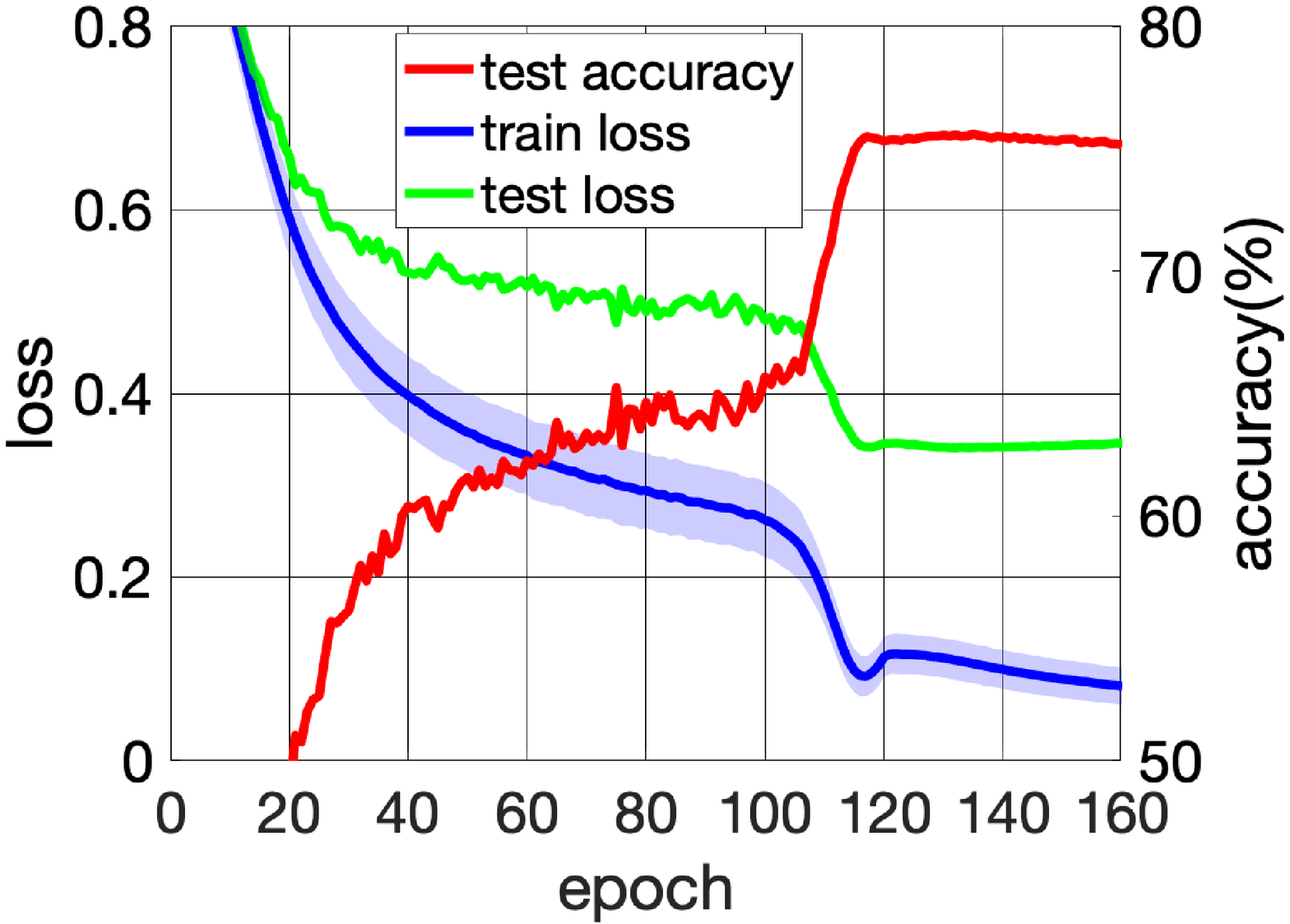}\\

\rotatebox[origin=c]{90}{SVHN} & 
\includegraphics[align=c,height=80pt]{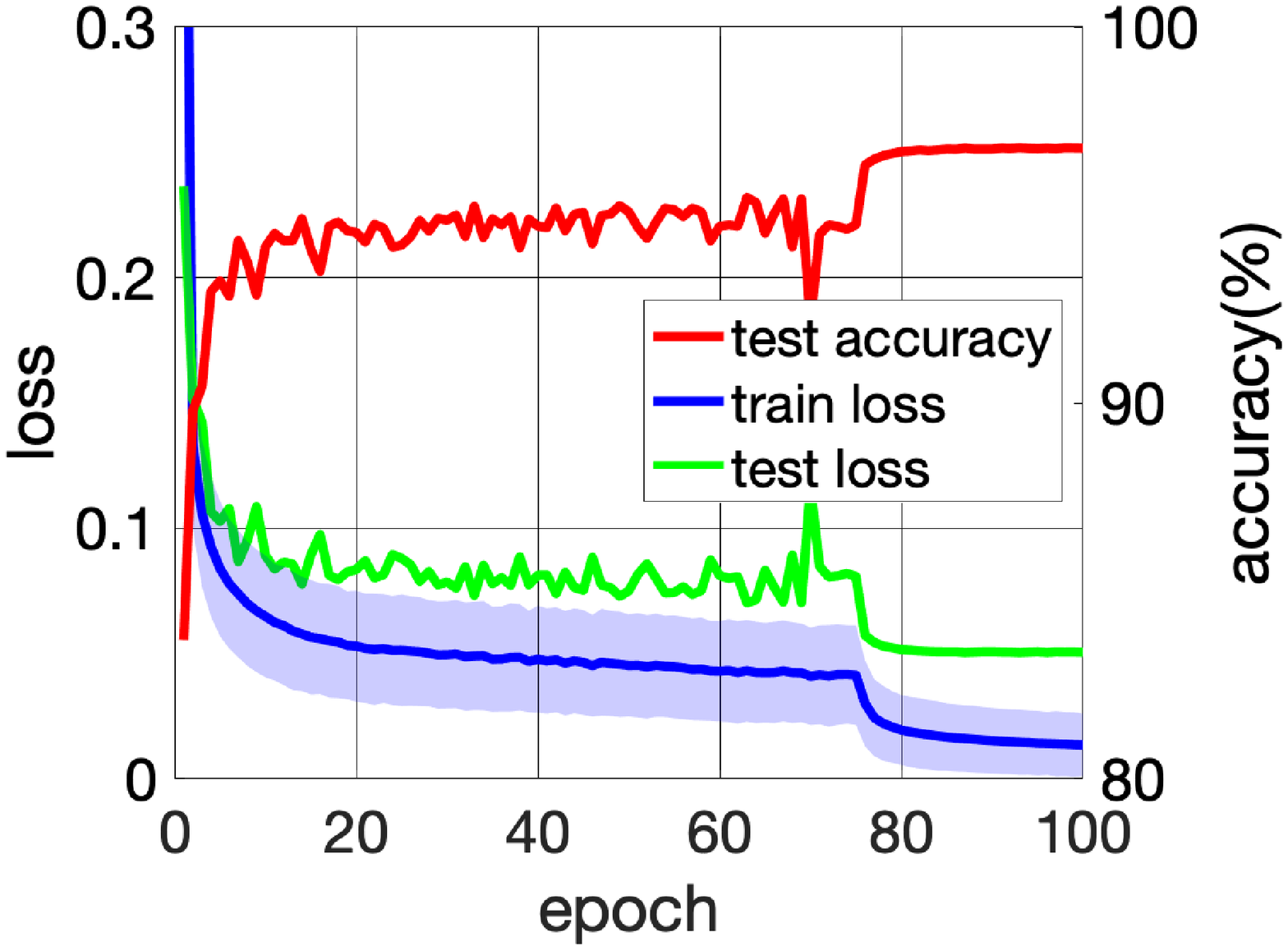} &
\includegraphics[align=c,height=80pt]{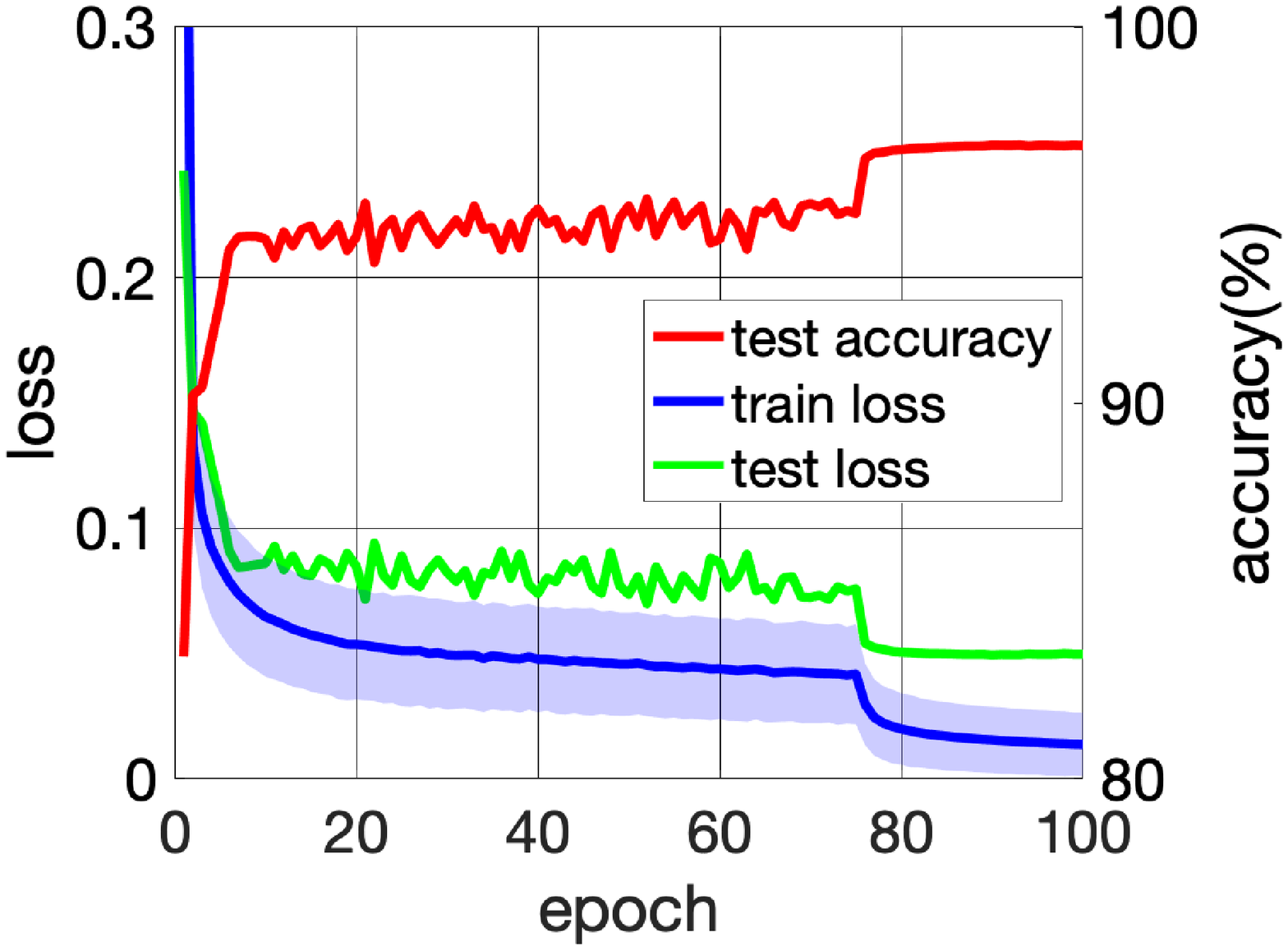} &
\includegraphics[align=c,height=80pt]{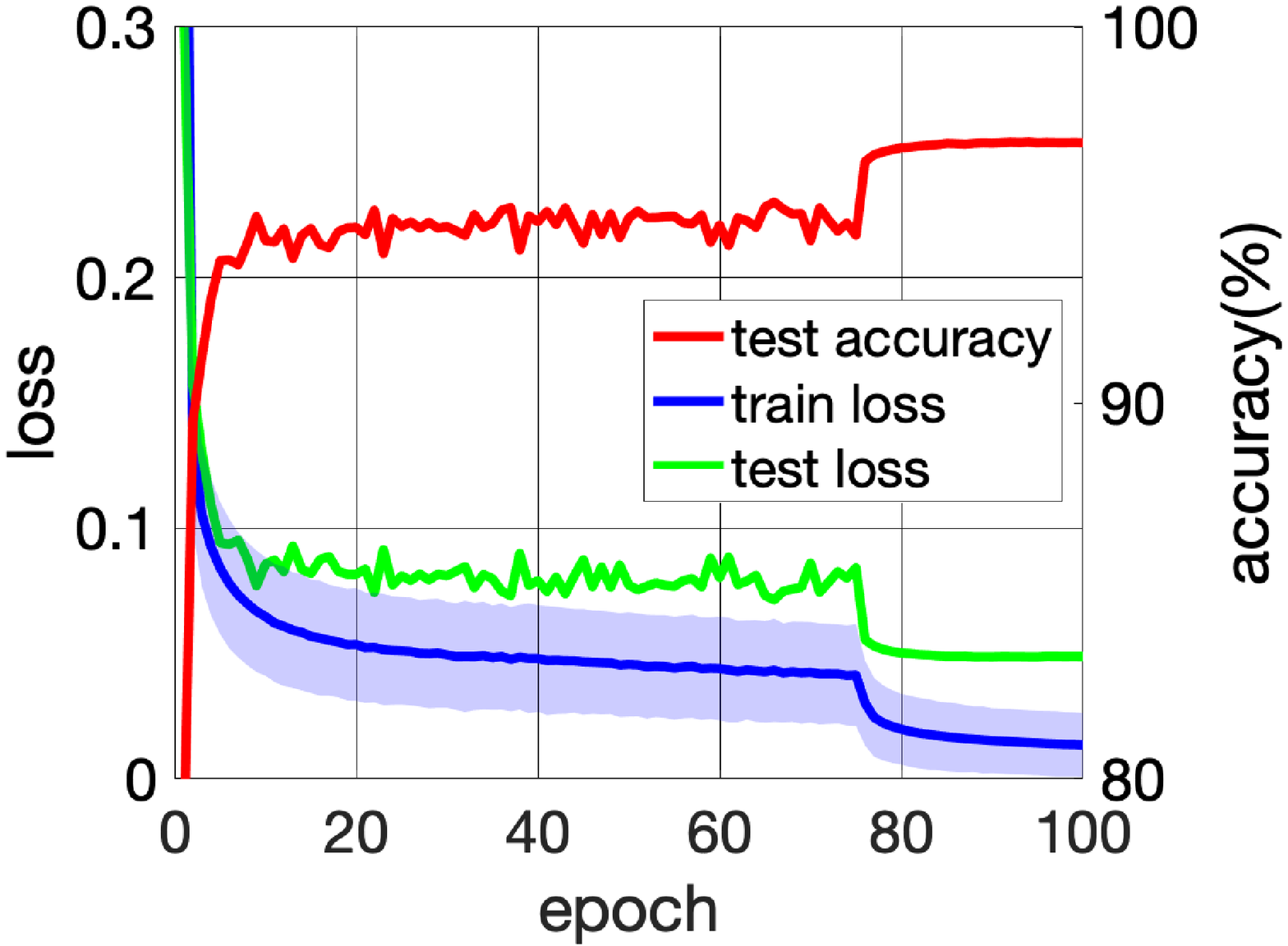}\\
& (a) SGD & (b) Laplace & (c) Logistic
\end{tabular}
\caption{Learning curves obtained based on DenseNet-BC ($k=12, L=100$) model using CIFAR-10 (top), CIFAR-100 (middle), and SVHN (bottom) datasets. The validation accuracy, training loss, testing loss are presented in red, blue, green color, respectively. The learning performance of our regularization scheme based on (b) Laplace and (c) Logistic distributions is compared with (a) the SGD algorithm.} 
\label{fig:densenet}
\end{figure*}

It is shown that our algorithm outperforms the other algorithms under comparison with their recommended parameters. Note that adaptive optimization methods such as Adam and AdaGrad often generalize worse than SGD for image classification task~\cite{wilson2017marginal}.
There is potential that our algorithm can be improved with wider range of grid search for the parameters $\alpha$ and $b$.
The learning curves obtained by (a) SGD and our algorithms with (b) Laplace and (c) Logistic distributions using CIFAR-10 (top), CIFAR-100 (middle), SVHN (bottom) are presented in Figure~\ref{fig:resnet56} and Figure~\ref{fig:densenet} where the learning curves with our algorithm indicate better generalization in terms of the validation accuracy. The learning rate is scheduled to be dropped at $75\%$ of epochs from $0.1$ to $0.001$ and the global adaptive annealing reaches the peak of the associated distribution at $75\%$ of epochs, which leads to an abrupt change in the learning curves.


%
%
%
\section{Conclusion and Discussion} \label{sec:conclusion}
In this paper, we have investigated the data-driven adaptive regularization by smoothing the residual of neural network for the image classification problem in an adaptive manner. The residual is defined by the discrepancy between the output of the neural network and the desired output. The regularization is imposed by diffusing the residual depending on the probability density function following either Laplace or Logistic distributions where the degree of regularization is proportional to the magnitude of each residual element. The combination of local and global annealing scheme that is designed to take into account residual in determining the degree of diffusion has been presented to spatially and temporally varying regularization.
The effectiveness of the proposed algorithm has been demonstrated by the experimental results indicating the potential of our algorithm that can be easily integrated to a variety of problems in deep learning applications.

\nolinenumbers

\bibliographystyle{spmpsci}      
\bibliography{image.bib}

\end{document}